\documentclass[preprint,12pt]{elsarticle}

\usepackage{amsmath,amssymb,amsfonts}
\usepackage{algorithmic}
\usepackage{graphicx}
\usepackage{textcomp}

\usepackage{makecell}
\usepackage{multirow}
\usepackage{multicol}
\usepackage{url}
\usepackage{textcomp}
\usepackage{color}
\usepackage{booktabs}
\usepackage{makecell}
\usepackage{subfigure}

\usepackage{tabularx}

\usepackage{float}

\usepackage{epsfig}

\usepackage{amssymb}
\usepackage{amsmath}

\journal{Medical Image Analysis}

\begin{document}

\begin{frontmatter}



\title{Concept Complement Bottleneck Model for Interpretable Medical Image Diagnosis} 


\author[a1]{Hongmei Wang} 
\ead{hwangfy@connnect.ust.hk} 

\author[a1]{Junlin Hou} 
\ead{csejlhou@ust.hk} 

\author[a1,a2]{Hao Chen\corref{cor1}} 
\ead{jhc@cse.ust.hk} 

\cortext[cor1]{Corresponding author}

\affiliation[a1]{organization={Department of Computer Science and Engineering, Hong Kong University of Science and Technology},
            city={Hong Kong},
            country={China}}
\affiliation[a2]{organization={Department of Chemical and Biological Engineering and Division of Life Science, Hong Kong University of Science and Technology},
            city={Hong Kong},
            country={China}}
\begin{abstract}
Models based on human-understandable concepts have received extensive attention to improve model interpretability for trustworthy artificial intelligence in the field of medical image analysis. These methods can provide convincing explanations for model decisions but heavily rely on the detailed annotation of pre-defined concepts. Consequently, they may not be effective in cases where concepts or annotations are incomplete or low quality. Although some methods automatically discover effective and new visual concepts rather than using pre-defined concepts or could find some human-understandable concepts via large language models, the discovered concepts are prone to veering away from medical diagnostic evidence or are challenging to understand. In this paper, we propose a concept complement bottleneck model for interpretable medical image diagnosis with the aim of complementing the existing concept set and finding new concepts to bridge the gap between explainable models and black-box models. Specifically, we propose to use concept adapters for specific concepts to mine the concept differences and score concepts in their own attention channels to support fairer concept learning. Then, we devise a concept complement strategy to learn new concepts while jointly using known concepts to improve model performance. Comprehensive experiments on medical datasets demonstrate that our model outperforms the state-of-the-art competitors in concept detection and disease diagnosis tasks while providing diverse explanations to ensure model interpretability effectively. 
\end{abstract}

\begin{graphicalabstract}
\includegraphics[scale=0.4]{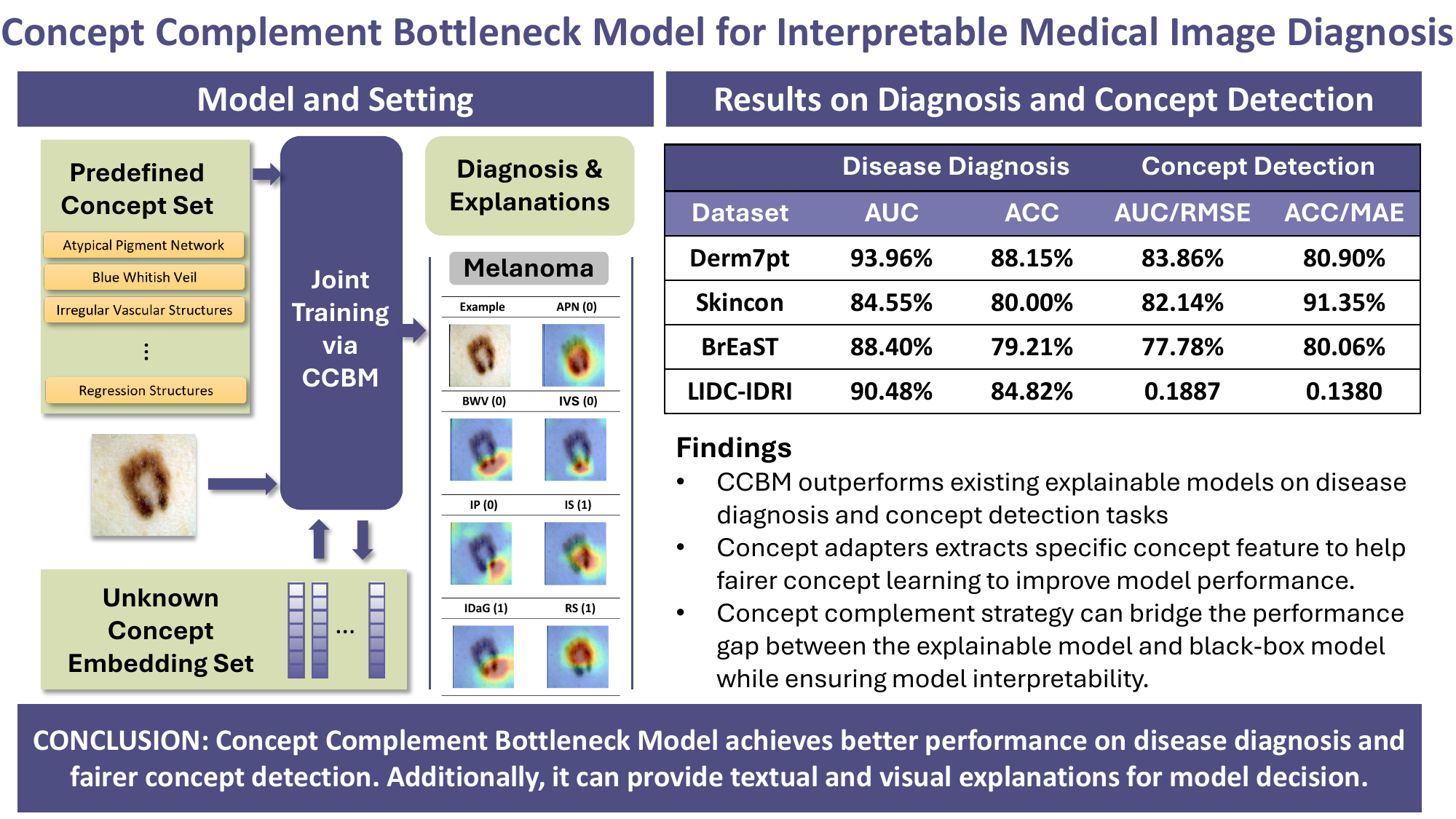}

\end{graphicalabstract}

\begin{highlights}
\item The proposed Concept Complement Bottleneck Model can leverage concept adapters and multi-head cross-attention to improve model performance on disease diagnosis and concept detection tasks.
\item Adding a concept adapter for each concept supports fairer concept learning so that the performance of concept detection is more balanced across different concepts.
\item Concept complement strategy can learn unknown concepts by jointly training with known concepts to help bridge the performance gap between explainable models and black-box models.
\item Extensive experiments demonstrate the effectiveness of concept complement bottleneck model in concept detection and disease diagnosis tasks while providing diverse visual and textual explanations for model decision.
\end{highlights}


\end{frontmatter}



\section{Introduction}

The progress of Artificial Intelligence (AI) has a profound impact on the development of image processing methods, where deep learning techniques provide excellent solutions for computer-aided diagnosis in Medical Image Analysis (MIA) such as pneumonia diagnosis \cite{asswin2023transfer, shea2023deep} and cancer detection \cite{wang2024weakly, luo2023deep}. Despite the impressive performance of deep learning models, they face various challenges in practical implementations. One crucial challenge is that black-box models lack transparency and interpretability during model end-to-end training. As for AI in MIA, it is crucial for doctors and patients to comprehend the significance and impact of data within the model's feature space in MIA, which allows them to place deeper trust in the model's predictions and support AI to further improve human well-being \cite{lucieri2020interpretability}.

Interpretable deep learning models can be divided into post-hoc models and ante-hoc models \cite{kasetty2024understanding}. Post-hoc models provide final explanations for model decisions after the model training process, such as gradient-based methods \cite{selvaraju2017grad, chattopadhay2018grad} and counterfactual-based methods \cite{cohen2021gifsplanation, singla2023explaining}. Although post-hoc explanations are expected to provide strong evidence for model decisions in many applications \cite{irvin2019chexpert, papanastasopoulos2020explainable}, their model decisions are sensitive to input data, indicating that the explanations are unreliable \cite{rudin2019stop}. Ante-hoc models ensure inherent interpretability throughout the entire process of model training and inference \cite{hou2024self}. In the field of MIA, ante-hoc explainable models mainly include attention-based \cite{li2021scouter, liu2024human}, example-based \cite{kim2021xprotonet, kong2023dp, hesse2024prototype} and concept-based models \cite{chauhan2023interactive, kim2023concept, patricio2023towards, hou2024concept}.

Concept-based ante-hoc explainable models have attracted widespread attention due to their ability to explain model decisions using human-friendly textual or visual concepts \cite{gupta2024survey}. Some studies focus on predicting concepts contained in the images and detailedly annotated by doctors or experts to make decision. For example, the Concept Bottleneck Model (CBM) \cite{koh2020concept} is one of the most representative works. CBM predicts concept scores by minimizing the concept classification cross-entropy loss. Then, it jointly or independently trains a classifier with a bottleneck layer to make decision in which step it allows externally adjusting the contribution of concepts to modify the final decision. Furthermore, many variants of CBM have been extensively studied in disease diagnosis based on X-ray, ultrasound and other medical images \cite{chauhan2023interactive, marcinkevivcs2024interpretable}. Although CBM-based methods effectively enhance model interpretability, they require fine-grained concept annotations of training data. However, in real-world scenarios, obtaining detailed annotations are extremely time-consuming and labor-consuming. There are some approaches that could automatically learn visual concepts \cite{fang2020concept, kong2024lce}, but the discovered concepts are difficult to correspond to clinical concepts in a general way. With the development of Large Language Models (LLMs) and Visual Language Models (VLMs), although many researches could automatically discover and generate textual concepts for images by using LLMs or VLMs \cite{shang2024incremental, yang2023language}, these concepts make more sense in the general field, but not so much in the medical field.

Despite that the mentioned methods have achieved good interpretability, they rely on annotated concepts completely or only use discovered concepts for decision-making. The former requires dense concept annotations and there is a performance gap between them and the black-box models \cite{hou2024self}, while the latter can easily deviate from real medical decision-making processes. Additionally, most of existing concept-based methods use the same image features for all concepts without considering the differences among concepts. On the visual level, simpler concepts are easier to capture, but in general, simple concepts contribute less to model decisions, especially for samples that are difficult to diagnose. Therefore, it is unfair to directly use the same feature encoded by an image encoder to calculate concept scores. 

To address these issues, we propose an interpretable model named Concept Complement Bottleneck Model (CCBM) for medical image disease diagnosis. It aims to achieve fair concept learning, keep the explanations from existing concept annotations and discover additional concepts that could collaborate with pre-defined concepts to help the model approach the performance of black-box models. Specifically, our contributions are as follows: 
1) Considering that one image encoder is hard to capture all concepts comprehensively, we configure a concept adapter for each concept to encode the most relevant part of the corresponding concept from the basic image feature to support fair concept learning and capture the regions most related to the concepts via Multi-Head Cross-Attention (MHCA). 2) We design a concept complement strategy to support unknown concept learning when jointly using known concepts, which could help the model approximate black-box performance by adding concept adapters and the concept aggregator directly. 3) CCBM can provide diverse visual and textual explanations for model decision. Especially, benefit from independent concept feature extracting and scoring, we can visualize all known or unknown concepts more effectively. 4) Extensive experiments demonstrate the effectiveness of our method in the aspects of concept detection as well as disease diagnosis. Model interpretability is verified by quantitative and qualitative experiments in the aspects of faithfulness, label effectiveness and understandability.

\section{Related Work}

Model interpretability is a crucial issue, especially in the field of MIA. The decisions of deep learning models in this high-risk area directly relate to human life and health. Therefore, for doctors and patients, any model decision must be reasonably explained, which could promote human trust model and make the next medical decision based on the model prediction. To enhance users' confidence in computer-aided or automated diagnostic systems, concept-based interpretable models for disease diagnosis have been studied widely. 

With the development of CBM, some research has been proposed to enhance the interpretability of disease diagnosis models based on the concept bottleneck. For instance, Yuksekgonul \textit{et al.} \cite{yuksekgonul2022pos} propose a Post-hoc Concept Bottleneck Model (PCBM) which flexibly obtains concept vectors from datasets with annotated concepts to generate concept subspace. Then, they map the feature of target domain data to the concept subspace to derive concept scores. Additionally, they also incorporate a residual network to enhance classification performance at the expense of model transparency. 
Chauhan \textit{et al.} \cite{chauhan2023interactive} introduce an interactive strategy that enables the model to seek input from human collaborators for labeling specific concepts, thereby enhancing the model's accuracy. Furthermore, some studies leverage textual concept encoding to aid in concept identification in images. Patricio \textit{et al.} \cite{patricio2023coherent} propose a two-stage skin disease diagnosis model. Initially, they derive lesion segmentation masks by a trained semantic segmentation model to masking invalid features. Subsequently, they append a concept encoding layer after the feature extractor to pinpoint the presence of specific concepts in the image and generate corresponding scores for classifier training. During training, they minimize the consistency loss between final visual concept encoding and textual concept encoding. Bie \textit{et al.} \cite{bie2024mica} present a skin disease diagnosis method based on multi-level alignment of image and textual concepts encoded by the pretrained LLM. They capture more detailed semantic information in the images through three levels of alignment: image-level, token-level, and concept-level. Similarly, they employ a multilayer perceptron with two fully connected layers (FCLs) to compute concept scores and diagnose skin diseases from aligned features. 

The methods mentioned above have greatly ameliorated the transparency of medical image diagnosis models, but they rely entirely on given medical concepts and fine-grained concept annotations. To reduce the dependence on concept annotations, some researchers present to discover image areas which could be defined as concepts \cite{fang2020concept, kong2024lce}. Fang \textit{et al.} \cite{fang2020concept} propose to clustering the regular image blocks which are effective for the model decision-making. Despite that the model could find the areas of the discover concepts are consistent with the doctors' explanations, the method is hard to be generalized to other diseases, like skin disease diagnosis due to the highly overlapping medical visual concepts. Additionally, some models have explored concept-based interpretable disease diagnosis methods using LLMs to generate concepts. For example, Patrício \textit{et al.} \cite{patricio2023towards} propose an embedded learning strategy to enhance the model's performance. They add trainable layers on top of the frozen CLIP to align categories with images. During inference, they predict image categories by calculating the similarity scores between text concepts generated by ChatGPT and the images. To further improve the semantic relevance between generated concepts and the input images, Kim \textit{et al.} \cite{kim2023concept} first generate candidate concepts using a LLM and then detect whether a given concept was present based on concept-based visual activation scores to remove invalid concepts. They only utilize visually meaningful concepts to guide and explain model decisions. However, all concepts share the same image features, which places higher demands on the feature extractor and limits the exploration of the different roles of various concepts in skincon disease diagnosis. Moreover, there still is the performance gap between black-box models and the existing methods based on fine-grained medical concepts and GPT-based concept generation. 

Hence, we design CCBM to ensure the collaboration between known and unknown concepts, which could keep the clinical interpretability of the model, reduce model reliance on annotations and while reducing the performance gap between our model and black-box models. 

\section{Methodology}

\begin{figure*}[t]
	\begin{center}
	  \includegraphics[width=1\linewidth]{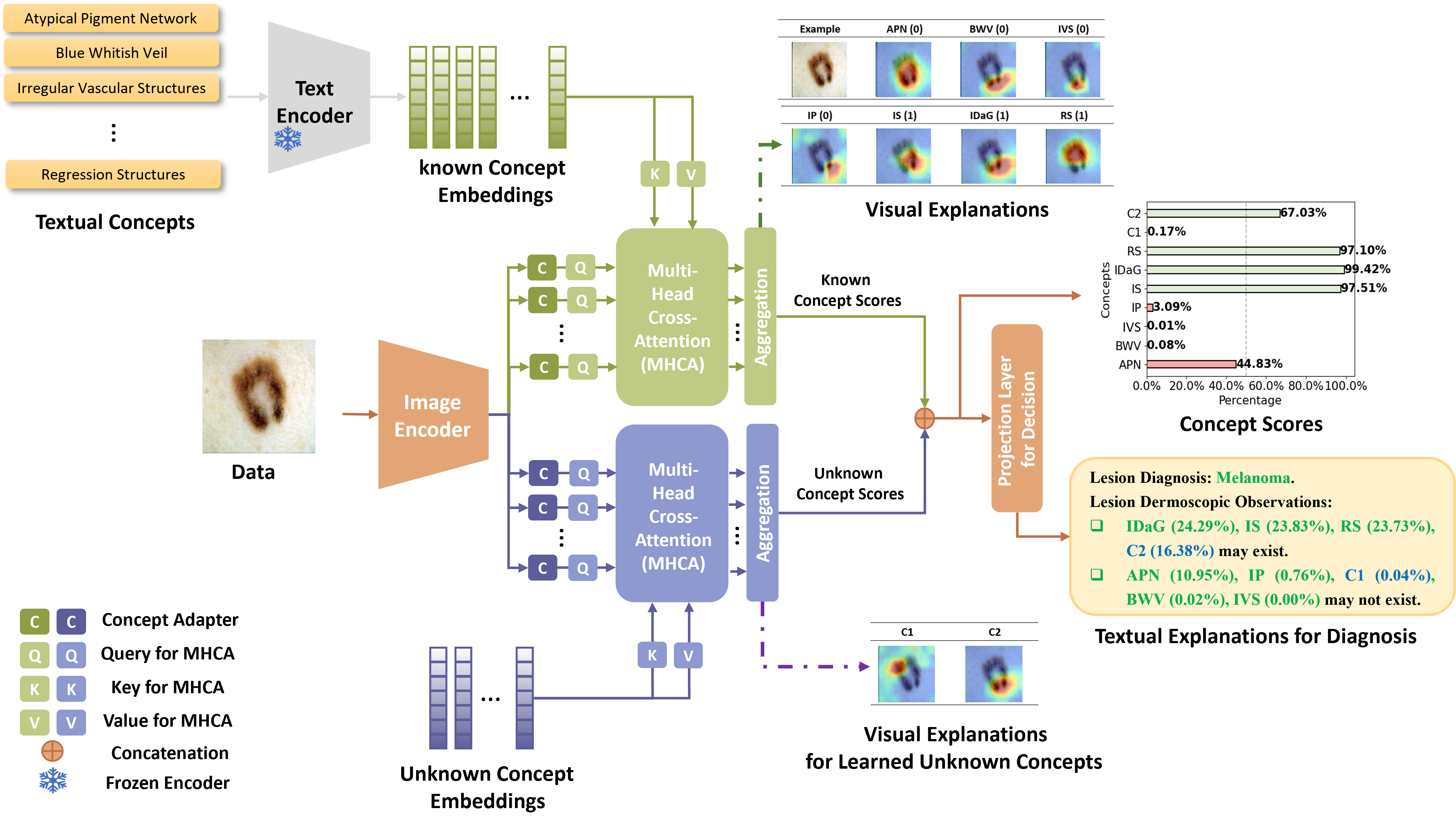}
	\end{center}
	\caption{\textbf{The Framework of Concept Complement Bottleneck Model}. The input images are delivered to the image encoder to obtain the fundamental features, then different concept adapters extract specific concept features. Next, CCBM calculates the visual-text cross-attention score between textual known concepts/unknown concept embeddings and concept visual features. Finally, these concepts attention scores are aggregated to be passed through the decision layer for final disease diagnosis.}
	\label{fig:framework}
\end{figure*}

In this paper, we present a Concept Complement Bottleneck Model (CCBM). Firstly, we utilize textual concepts to guide the model to learn high-dimensional medical concepts. Secondly, by configuring the concept adapters, the model extracts features that are conducive to learning specific concepts and aggregates concept scores from cross-attention modules. Finally, we propose a concept complement strategy to learn new and effective concepts, which ensures the interpretability of the model while reducing the performance gap between the interpretable model and the black-box model. The overall framework of the model is shown in Fig. \ref{fig:framework}. We construct CCBM by the following steps below.

\subsection{Textual and Visual Concept Feature Extraction}

We first extract the textual known concept embeddings and then specifically learn visual features through the concept adapters in this step. 

\subsubsection{Encoding Textual Known Concepts}
The textual known concept set is defined as $\mathcal{K}$ which includes $n_k$ labeled concepts, and the textual known concept encoder $E_T$ is used to encode the textual known concepts. In our settings, $E_T$ is a frozen text encoder to map the concepts to $d_k$-dimensional concept subspace: $E_{T}: \mathcal{K} \rightarrow \mathbb{R}^{d_k}$. Furthermore, these known concept embeddings will be used as the keys $K_i$ $(i=1,2,...,n_k)$ and values $V_i$ $(i=1,2,...,n_k)$ in the cross-attention module. 

\subsubsection{Encoding Visual Known Concepts Independently}

We set up concept adapters to extract key features from a shared image encoding for each concept, considering their unique differences. Detailedly, we select a Convolutional Neural Network (CNN) $E_{I}$ as the image encoder to map the input image data $\mathcal{X}$ including $n$ samples from $n_c$ classes to the fundamental feature space: $E_{I}:$ $\mathcal{X} \rightarrow \mathbb{R}^d$, where $d$ represents the feature dimension. After that, we set $n_k$ concept adapters $C_i$ $(i=1,2,...,n_k): \mathbb{R}^d \rightarrow \mathbb{R}^{d_k}$ to extract specific concept features from the fundamental features to be the queries $Q$ of the following cross-attention module. Intuitively, we hope the fundamental features extracted by the image encoder can include all of crucial features for all different concepts, so we had better to choose a strong image encoder. For simplicity, in our setting, each concept adapter is set as a FCL that maps the $d$-dimension fundamental feature space to the ${d_k}$-dimension concept subspace. It is worth noting that we need to ensure that the subspace dimensions of all concept adapters are consistent with the subspace dimension of text encoder. We fomulate the image encoder and concept adapters as follows:
\begin{equation}
	Q_i = C_i( E_{I}(\mathcal{X})), i=1,2,...,n_k,
\end{equation}
in which $Q_i$ is the specific concept features extracted by the $i$-th concept adapter.

\subsection{Concept Score Prediction} 
Through the encoders and concept adapters, we can obtain the image features of all known concepts and the embedings of all textual known concepts. In this step, we use the multi-head cross-attention module to obtain the attention score. For all known concepts, we use the embeddings of the textual concept as the keys and values, and the outputs of the concept adapters as the queries to calculate the attention weights where each output is a query. All textual concept embeddings are represented as the matrix $K \in \mathbb{R}^{n_k \times d_k}$ and $V \in \mathbb{R}^{n_k \times d_k}$, and all visual concept embeddings are denoted as the matrix $Q \in \mathbb{R}^{n_k \times d_k} $. In this paper, we only fomulate the expression of single-head cross-attention output: 
\begin{align}
  A (Q, K) = softmax\left(\frac{QK^T}{\sqrt{d_k}}\right),\\
  A_w (Q,K,V) = A(Q,K)V,
\end{align}
where $A \in \mathbb{R}^{n_k \times n_k} $ is the attention map matrix whose elements represent the weights of different concept pairs and $A_w \in \mathbb{R}^{n_k \times d_k}$ is the weighted sum of all concepts. According to the setting of the concept adapters, we can not average these attention as final features to calculate concept scores by a FCL but need to aggregate them in another way to get concept scores independently. Specifically, we can get the attention $A_w(Q_i, K, V)$ for the $i$-th concept. Furthermore, we need to calculate the concept scores based on these attention weights. Different from the previous bottleneck models which directly use a FCL to project the common feature to get the concept scores, we could calculate these concept scores independently by any score calculation module based on their specific concept features. In our model, the aggregator for known concepts includes $n_k$ FCLs, and we use each FCL $f_{i}$ to project the attention weights to get the concept score for each concept. Hence, the concept scores are calculated as follows:
\begin{equation}
	S_i = f_i(A_w(Q_i, K, V)), i=1,2,...,n_k,
\end{equation}
where $S_i$ is the final concept score for the $i$-th concept.

\subsection{Concept Complement Bottleneck Model for Medcial Image Diagnosis}
\subsubsection{Explainable Diagnosis Decision using Pre-defined Concept Set}
If we do not set unknown concept learning branch in our model, based on these concept scores, we can directly predict the diagnosis results by a decision layer $f_d$. The final prediction is:
\begin{equation}
	\hat{Y} = f_d(S) \in \mathbb{R}^{n_c},
\end{equation}
where $S=[S_1,S_2,...,S_{n_k}]$ is the concept score vectors of input images, $n_c$ is the number of classes and $\hat{Y}$ is the final disease prediction. 

During the training process, we jointly train the model to perform well on the concept detection task and disease diagnosis task. In particular, we require model decisions to more explicitly depend on these concept scores to ensure model interpretability. Therefore, we leverage the cross-entropy loss for the classification task and the concept-learning loss for the concept detection task. The total loss of our CCBM is:
\begin{equation}
	\min_{\hat{Y}, S} \left( \lambda_1 \mathcal{L}_{ce}(\hat{Y}, Y) + \mathcal{L}_{cep}(S, C) \right),
\end{equation}
where $\hat{Y}$ is the classification prediction of the model, $Y$ is the ground truth of image diagnosis, $C$ is the matrix of the ground truth of the concept detection task, and $\lambda_1$ is the hyperparameter to balance the two tasks. $\mathcal{L}_{ce}$ is the cross-entropy loss for classification task:
\begin{equation}
	\mathcal{L}_{ce} = -\sum_{i=1}^{n} \sum_{j=1}^{n_c} y_{ij} \log(\hat{y}_{ij}),
\end{equation}
and $\mathcal{L}_{cep}$ depends on the concept learning task is a classification task or a regression task. As for the former, we use the multi-label classification cross-entropy loss, which can be formulated as:
\begin{equation}
	\mathcal{L}_{cep}= -\sum_{i=1}^{n} \sum_{j=1}^{n_k} \left( c_{ij} \log(s_{ij}') + (1 - c_{ij}) \log(1 - s_{ij}') \right),
\end{equation}
and as for the latter, if the concept annotation is scored but not binary, we use the Mean Square Error (MSE) loss $\mathcal{L}_{mse}^{cep}$:
\begin{equation}
	\mathcal{L}_{cep}= \frac{1}{n}\sum_{i=1}^{n} \sum_{j=1}^{n_k} (c_{ij} - s_{ij})^2,
\end{equation}
where $s_{ij}$ and $c_{ij}$ is the concept label and the ground truth of the $j$-th concept of the $i$-th sample, respectively. $s_{ij}' = \sigma(s_{ij})$, and $\sigma(\cdot)$ is the sigmoid function. 

\subsubsection{Concept Complement Strategy}

Although existing models based on concept bottleneck can maintain high classification performance while providing interpretability by training on concept detection tasks and classification tasks, the model performance needs to be balanced between concept detection and classification tasks. In addition, there is still a gap between explainable models and black-box models in classification tasks. In order to reduce this performance gap while maintaining model transparency, we propose a concept complement strategy to learn unknown concepts which are helpful to diagnosis.

The concept complement strategy includes two parts: concept complement and concept score adjustment. Concept complement means that we add $n_u$ additional unknown concept adapters which are set as $n_u$ FCLs projecting features from $d$-dimension to $d_u$-dimension space to learn $n_u$ new concepts. The general processing of unknown concepts is similar to that of known concepts in the other steps. However, we have no textual representions of these new concepts. We set $n_u$ learnable embedding vectors in the model to uniquely represent the learned unknown concepts which will be used as the keys $K_j^u$ and values $V_j^u$ of unknown concept cross-attention module. If we denote the visual encoding of unknown concepts from concept adapters as $Q_{j}^{u} \in \mathbb{R}^{n_u \times d_{u}}$, the concept scores of unknown concepts are calculated as:
\begin{equation}
	l_{j} = g_j(A_w(Q_{j}^{u}, K^u, V^u)) \in \mathbb{R}^{n_{u}}, j=1,2,...,n_{u},
\end{equation}
where $l_{j}$ is the final concept score for the $j$-th unknown concepts and $g_j$ is the FCL for aggregating the $j$-th unknown concept. It is worth noting that these scores are only used to help adjust the importance of known concepts so that the model can better approach the performance of the black-box model, since we have no label of these concepts. The final diagnosis of model can be represented as:
\begin{equation}
	\hat{Y} = f'_d([S,L]) \in \mathbb{R}^{n_c},
\end{equation}
where $L=[l_1,l_2,...,l_{n_u}]$ is the concept scores of unknown concepts, $[S,L]$ is the concatenation of the concept scores of known and unknown concepts, and $f'_d$ is the FCL for the final prediction. Additionally, we add a similarity loss term to the total loss to ensure the unknown concepts are as different as possible from the known concepts and as different as each other. The total loss of CCBM with concept complement is:
\begin{align}
	\nonumber
  \min_{\hat{Y}, S, L} & \Bigg( \lambda_1 \mathcal{L}_{ce}(\hat{Y}, Y) + \mathcal{L}_{cep}(S, C) \\
  + &\lambda_2 \sum_{i=1}^{n_u} (\sum_{j \neq i}^{n_u} \mathcal{L}_{sim}(K_i^u, K_j^u) + \sum_{j =1}^{n_k} \mathcal{L}_{sim}(K_i^u, K_j) ) \Bigg),
\end{align}
where $\mathcal{L}_{sim}$ is the cosine similarity loss for concept learning.

\section{Experiments}
In this section, we introduce our experiment datasets, settings and comparison algorithms, and we report and analyze our experimental results to verify the effectiveness of CCBM. 

We conducted comprehensive experiments on two dermoscopic image datasets \textit{Derm7pt} and \textit{Skincon}, an ultrasound breast image dataset \textit{BrEaST}, and a lung nodule CT dataset \textit{LIDC-IDRI}. 
\textbf{Derm7pt} \cite{kawahara2018seven} contains 1011 dermoscopic images including 20 specific skin disease diagnosis and detailed labels of 7 clinical concepts based on the seven-point skin lesion malignancy checklist. We only considered 827 images in which the diagnosis belongs to \textit{Nevus} and \textit{Melanoma}.
\textbf{Skincon} \cite{daneshjou2022skincon} contains 3230 skin images of \textit{Malignant}, \textit{Benign}, and \textit{Non-neoplastic} categories in the Fitzpatrick 17k dataset \cite{groh2021evaluating}. We choose 22 concepts where there are at least 50 images containing the concept from 48 general medical clinical concepts densely annotated by two dermatologists.
\textbf{BrEaST} \cite{pawlowska2024curated} is an ultrasound breast image dataset with detailed annotations via 7 concepts from BI-RADS descriptors, which contains 256 images with 3 different types of breast diagnosis, including \textit{Benign}, \textit{Malignant} and \textit{Normal}. We only use 254 abnormal breast images in our experiments. 
\textbf{LIDC-IDRI} \cite{armato2015data} contains 1018 CT scans with detailed annotations for 8 concepts from experienced radiologists. We extract the regions of interest of 2635 lung nodules with a diameter over 3 mm from 2D CT images for experiments. To address the annotation disagreement among the radiologists, we calculate the average of the benign-malignant ranking as the malignant score for the disease diagnosis task. Then, we assign binary labels to nodules based on their 0-5 averaged malignancy score. If the malignancy score is over 3, we define the nodule as \textit{Malignant}; otherwise, it is \textit{Benign}. Similarly, for each concept, we normalized scores to a range of 0 to 1 and averaged the different scores from radiologists to obtain the final concept scores. The used concepts are detailedly shown in Table \ref{tab:dataset}.

\begin{table}[t]
	\centering
	\caption{Concepts in Our Experiments. } 
  \scriptsize
	\begin{tabular}{cccccc}
	\toprule
	Dataset &  Used Concepts  \\
	\midrule
	Derm7pt  & \makecell{Atypical Pigment Network (APN), Blue Whitish Veil (BWV), \\Irregular Vascular Structures (IVS), Irregular Pigmentation(IP), Irregular Streaks (IS), \\ Irregular Dots and Globules (IDaD), Regression Structures (RS)}  \\
	\midrule
	Skincon  & \makecell{ PAPule (PAP), PLAque (PLA),PUStule (PUS), BULla (BUL),\\ PATch (PAT), NODule (NOD), ULCer (ULC), CRUst (CRU),\\ EROsion (ERO), ATRophy (ATR), EXUdate (EXU), TELangiectasia (TEL), \\SCALe (SCAL),  SCAR (SCAR), FRIable (FRI),Dome-SHaped (DSH), \\ Brown-Hyperpigmentation (BrH), White-Hypopigmentation (WhH), PURple (PUR), \\ YELlow (YEL), BLAck (BLA), ERYthema (ERY)} \\
	\midrule
	BrEaST  & \makecell{Irregular SHape (IRS), Not Circumscribed Margin (NCM), \\ Hyperechoic or Heterogeneous Echogenicity (HoHE),\\ Posterior Features (PF), Hyperechoic Halo (HH), \\ CALcifications (CAL), Skin Thickening (ST)}\\
	\midrule
	LIDC-IDRI  & \makecell{SUBtlety (SUB), InternalSTructure (IST), \\ CALcification (CAL), SPHericity (SPH), MARgin (MAR), \\ LOBulation (LOB), SPIculation (SPI), TEXture (TEX)}\\
	\bottomrule
	\end{tabular}
	\label{tab:dataset}
\end{table}	

\subsection{Implementation Details}
As for dermoscopic image datasets \textit{Derm7pt} and \textit{Skincon}, we follow PCBM \cite{yuksekgonul2022pos} and choose the trained Inception-v3 model \cite{szegedy2015going} as the image encoder, and we use pretrained ResNet50 as the image encoder for other two datasets. The frozen pre-trained ClinicalBERT \cite{alsentzer2019publicly} is utilized as the text encoder. 
In our experiments, the concept adapters and aggregators are set as FCLs. 
Additionally, we train the model for 300 epochs at most using the Adam optimizer and we stop the training process if there is no significant change in training loss. We use the grid method to choose model hyperparameters. For the \textit{Derm7pt} and \textit{BrEaST} dataset, the hyperparameters are set as $\lambda_1=0.2, \lambda_2=10$. For the \textit{Skincon} dataset, $\lambda_1=0.1, \lambda_2=5$, and $\lambda_1=0.5, \lambda_2=10$ for \textit{LIDC-IDRI} dataset. As for the number of unknown concepts, we can set $n_{u}$ as the number of categories of datasets, since intuitively we need at least the same number of data features as the sample categories to distinguish all categories. To evaluate model performance, we use the Area Under Curve (AUC), ACCuracy (ACC) and F1-score as disease diagnosis evaluation metrics, and the first two metrics are also used to evaluate the concept detection tasks of \textit{Derm7pt}, \textit{Skincon} and \textit{BrEaST}. Root Mean Square Error (RMSE) and Mean Absolute Error (MAE) are used for evaluating the concept Regression task of \textit{LDIC-IDRI} dataset. All of mean and standard deviation results are obtained by five-fold cross-validation in our experiments. 

\subsection{Comparison Algorithms}
To verify the effectiveness and advancement of the model, we compare CCBM with the state-of-the-art methods, including CBM \cite{koh2020concept}, PCBM (-H) \cite{yuksekgonul2022pos}, an Ante-hoc Explainable Concept-based model (AEC) \cite{sarkar2022framework}, Concept-Based Interpretability using Vision-Language Models (CBIVLM) \cite{patricio2023towards} and Energy-based CBM (ECBM) \cite{xu2024energy}. We also test the backbone models to evaluate the gap between our explainable model and black-box models. For methods that do not support concept detection, we use ``N/A'' to indicate invalid data.

\subsection{Experimental Results and Analysis}

\subsubsection{Disease Diagnosis and Concept Detection}

\begin{table}[t]
	\centering
	\caption{Quantitative results on disease diagnosis and concept detection tasks with comparison methods and black-box models. The results are shown as the mean $\pm$ std of five-fold cross-validation experiment. } 
  \scriptsize
	\setlength{\tabcolsep}{1pt} 
	\begin{tabular}{ccccccccccc}
	\toprule
	\multicolumn{1}{c}{\multirow{2}{*}{Dataset}} & \multicolumn{1}{c}{\multirow{2}{*}{Model}} & \multicolumn{3}{c}{Disease Diagnosis} & \multicolumn{2}{c}{Concept Detection} \\
	\cmidrule(lr){3-5} \cmidrule(lr){6-7}
	& & AUC (\%) $\uparrow $ & ACC (\%) $\uparrow $ & F1-score (\%) $\uparrow $ & AUC (\%) $\uparrow $ & ACC (\%) $\uparrow $ \\
	\midrule
	\multicolumn{1}{c}{\multirow{7}{*}{\textit{Derm7pt}}} 
	& PCBM \cite{yuksekgonul2022pos} & 81.32 $\pm$ 2.12 & 75.82 $\pm$ 2.00 & 71.10 $\pm$ 1.74 & N/A & N/A \\
	& PCBM-H \cite{yuksekgonul2022pos} & 85.87 $\pm$ 1.53 & 78.60 $\pm$ 2.79 & 75.50 $\pm$ 3.09 & N/A & N/A \\
	& CBIVLM \cite{patricio2023towards} & 83.45 $\pm$ 3.59 & 84.13 $\pm$ 2.71 & 71.24 $\pm$ 2.54 & N/A & N/A \\
	& ECBM \cite{xu2024energy} & 77.44 $\pm$ 1.84 & 79.69 $\pm$ 1.84 & 76.57 $\pm$ 0.00 & 69.54 $\pm$ 1.79 & 77.73 $\pm$ 1.47 \\
	& AEC \cite{sarkar2022framework} & 91.27 $\pm$ 2.02 & 84.88 $\pm$ 2.05 & 80.99 $\pm$ 3.32 & 76.61 $\pm$ 1.61 & 75.30 $\pm$ 0.72 \\
	& CBM \cite{koh2020concept} & 92.88 $\pm$ 1.90 & 85.89 $\pm$ 1.92 & 82.18 $\pm$ 2.67 & 82.15 $\pm$ 2.68 & 80.00 $\pm$ 1.87 \\
	&CCBM & \textbf{93.96} $\pm$ \textbf{0.95} & \textbf{88.15} $\pm$ \textbf{1.64} & \textbf{85.61} $\pm$ \textbf{2.79} & \textbf{83.86} $\pm$ \textbf{1.00} & \textbf{80.90} $\pm$ \textbf{1.56} \\
	\cmidrule{2-7}
	& Inception-v3 & 92.02 $\pm$ 2.53 & 86.46 $\pm$ 2.54 & 83.13 $\pm$ 3.31 & N/A & N/A \\
	\midrule
	\multicolumn{1}{c}{\multirow{7}{*}{\textit{Skincon}}} 
	& PCBM \cite{yuksekgonul2022pos} & 69.06 $\pm$ 1.23 & 72.48 $\pm$ 1.56 & 39.55 $\pm$ 0.97 & N/A & N/A \\
	& PCBM-H \cite{yuksekgonul2022pos} & 72.85 $\pm$ 1.66 & 68.42 $\pm$ 3.07 & 53.33 $\pm$ 3.38 & N/A & N/A \\
	& CBIVLM \cite{patricio2023towards} & 71.34 $\pm$ 2.16 & 70.74 $\pm$ 1.11 & 66.15 $\pm$ 1.50 & N/A & N/A \\
	& ECBM \cite{xu2024energy} & 68.72 $\pm$ 1.73 & 78.79 $\pm$ 1.41 & 61.12 $\pm$ 0.00 & 65.11 $\pm$ 0.89 & 90.94 $\pm$ 0.25 \\
	& AEC \cite{sarkar2022framework} & 83.86 $\pm$ 0.61 & \textbf{80.24} $\pm$ \textbf{0.52} & 63.89 $\pm$ 1.63 & 58.64 $\pm$ 0.90 & 90.64 $\pm$ 0.10 \\
	& CBM \cite{koh2020concept} & 80.01 $\pm$ 1.25 & 78.42 $\pm$ 1.31 & 60.57 $\pm$ 3.04 & 62.14 $\pm$ 1.37 & 89.32 $\pm$ 0.14 \\
	&CCBM & \textbf{84.55} $\pm$ \textbf{1.87} & 80.00 $\pm$ 1.75 & \textbf{67.14} $\pm$ \textbf{1.87} & \textbf{82.14} $\pm$ \textbf{0.33} & \textbf{91.35} $\pm$ \textbf{0.25} \\
	\cmidrule{2-7}
	& Inception-v3 & 79.92 $\pm$ 1.48 & 77.52 $\pm$ 1.47 & 59.86 $\pm$ 2.78 & N/A & N/A \\
	\midrule
	\multicolumn{1}{c}{\multirow{6}{*}{\textit{BrEaST}}} 
	& PCBM \cite{yuksekgonul2022pos} & 75.41 $\pm$ 5.74 & 68.43 $\pm$ 7.64 & 64.79 $\pm$ 8.28 & N/A & N/A \\
	& PCBM-H \cite{yuksekgonul2022pos} & 79.63 $\pm$ 2.95 & 70.02 $\pm$ 3.87 & 67.73 $\pm$ 3.81 & N/A & N/A \\
	& ECBM \cite{xu2024energy} & 71.38 $\pm$ 3.84 & 73.63 $\pm$ 5.28 & 71.33 $\pm$ 0.00 & 62.50 $\pm$ 2.67 & 79.94 $\pm$ 1.79 \\
	& AEC \cite{sarkar2022framework} & 82.80 $\pm$ 3.32 & 72.40 $\pm$ 3.39 & 68.60 $\pm$ 3.51 & 69.71 $\pm$ 1.52 & 77.82 $\pm$ 1.70 \\
	& CBM \cite{koh2020concept} & 87.42 $\pm$ 4.27 & 77.21 $\pm$ 8.62 & 76.29 $\pm$ 8.31 & 70.76 $\pm$ 1.38 & 77.49 $\pm$ 1.43 \\

	&CCBM & \textbf{88.49} $\pm$ \textbf{5.85} & \textbf{79.21} $\pm$ \textbf{6.00} & \textbf{77.95} $\pm$ \textbf{5.53} & \textbf{77.78} $\pm$ \textbf{4.00} & \textbf{80.06} $\pm$ \textbf{3.46} \\

	\cmidrule{2-7}
	& ResNet50 & 86.97 $\pm$ 6.14 & 77.61 $\pm$ 6.23 & 76.39 $\pm$ 6.09 & N/A & N/A \\
	\midrule
	& & AUC (\%) $\uparrow $ & ACC (\%) $\uparrow $ & F1-score (\%) $\uparrow $ & RMSE $\downarrow $ & MAE $\downarrow $\\
	\midrule
	\multicolumn{1}{c}{\multirow{6}{*}{\textit{LIDC-IDRI}}} 
	& PCBM \cite{yuksekgonul2022pos} & 85.09 $\pm$ 1.33 & 81.47 $\pm$ 1.24 & 77.96 $\pm$ 1.59 & N/A & N/A \\
	& PCBM-H \cite{yuksekgonul2022pos} & 88.77 $\pm$ 1.64 & 83.28 $\pm$ 1.57 & 80.71 $\pm$ 1.46 & N/A & N/A \\
	& AEC \cite{sarkar2022framework} & 87.35 $\pm$ 2.16 & 84.23 $\pm$ 1.67 & 81.57 $\pm$ 1.86 & 0.5710 $\pm$ 0.0100 & 0.5064 $\pm$ 0.0094 \\
	& CBM \cite{koh2020concept} & 89.76 $\pm$ 1.30 & 83.63 $\pm$ 1.67 & 80.22 $\pm$ 1.85 & 0.2981 $\pm$ 0.0096 & 0.2461 $\pm$ 0.0095 \\
	&CCBM & \textbf{90.48} $\pm$ \textbf{0.94} & \textbf{84.82} $\pm$ \textbf{0.43} & \textbf{82.56} $\pm$ \textbf{0.67} & \textbf{0.1890} $\pm$ \textbf{0.0046} & \textbf{0.1380} $\pm$ \textbf{0.0032}\\
	\cmidrule{2-7}
	& ResNet50 & 90.18 $\pm$ 1.17 & 84.08 $\pm$ 1.71 & 81.70 $\pm$ 1.52 & N/A & N/A \\
	\bottomrule
	\end{tabular}
	\label{tab:classification}
	\end{table}	

To verify the effectiveness of our model on concept detection task and skin disease diagnosis task, we conducted extensive experiments and compared CCBM with the state-of-the-art methods on two datasets using multiple metrics. The five-fold cross validation results of CCBM on two tasks are shown in Table \ref{tab:classification}. 

For the \textit{Derm7pt} dataset, CCBM achieves the best performance in the concept detection task, significantly outperforming the comparison explainable methods with 93.96\% AUC, 88.15\% ACC and 85.61\% F1-scores. Additionally, CCBM surpasses the black-box model by 1\% to 2\% on all classification task metrics. Regarding the \textit{Skincon} dataset, CCBM excels in the concept detection task, particularly with the AUC 20\% higher than other competitors, and the best ACC of 91.35\%. While maintaining strong concept detection performance, the AUC reaches 84.55\%, outperforming other competitors, and the ACC achieves an impressive 80.00\%, slightly lower than the best result of 80.24\%. The black-box model underperforms on the dataset, with CCBM surpassing it by 5\% on AUC, 3\% on AUC and 7\% on F1-score. 

The results of the \textit{BrEaST} dataset show that CCBM achieves the best performance on the concept detection task and the classification task on all evaluation metrics. The concept detection AUC is 77.78\% which is 7\% higher than CBM. The classification AUC, ACC and F1-score are 88.49\%, 79.21\% and 77.95\%, respectively, representing an improvement of about 1\% to 2\% over CBM. In comparison to the black-box model, CCBM outperforms it by approximately 1.5\% across all metrics. CCBM also demonstrates superior performance in both the concept regression task and the classification task across all evaluation metrics on the \textit{LIDC-IDRI} dataset. For classification task, the results are also 1\% to 2\% higher than CBM. Additionally, while CCBM clearly outperforms the black-box model in the classification task, the performance gap is not as pronounced as in the other datasets. 

Overall, since CCBM learns the unknown concepts to complement the known concepts which provide additional information for the model decision, it outperforms other advanced explainable models on disease diagnosis while guaranteeing the best performance on concept detection, and it also exceeds the black-box model on the classification task. 

\begin{figure*}[t]
  \centering
  \subfigure[\textit{Concept AUC on Derm7pt}]{
    \begin{minipage}[b]{0.45\textwidth}
      \includegraphics[width=1\textwidth]{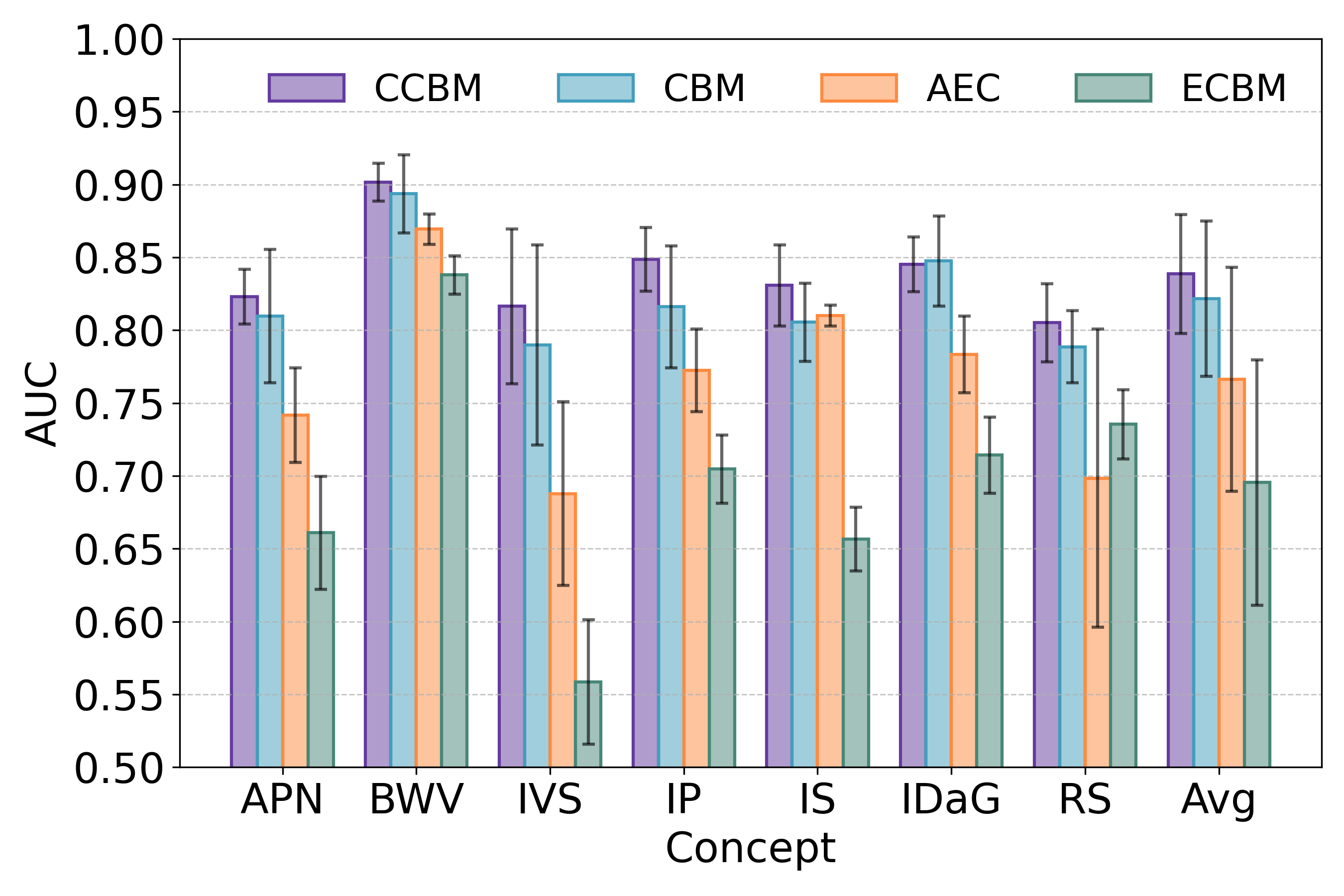}
    \end{minipage}
    \label{fig:cep_auc_derm7pt}
  } 
  \subfigure[\textit{Concept AUC on BrEaST}]{
    \begin{minipage}[b]{0.45\textwidth}
      \includegraphics[width=1\textwidth]{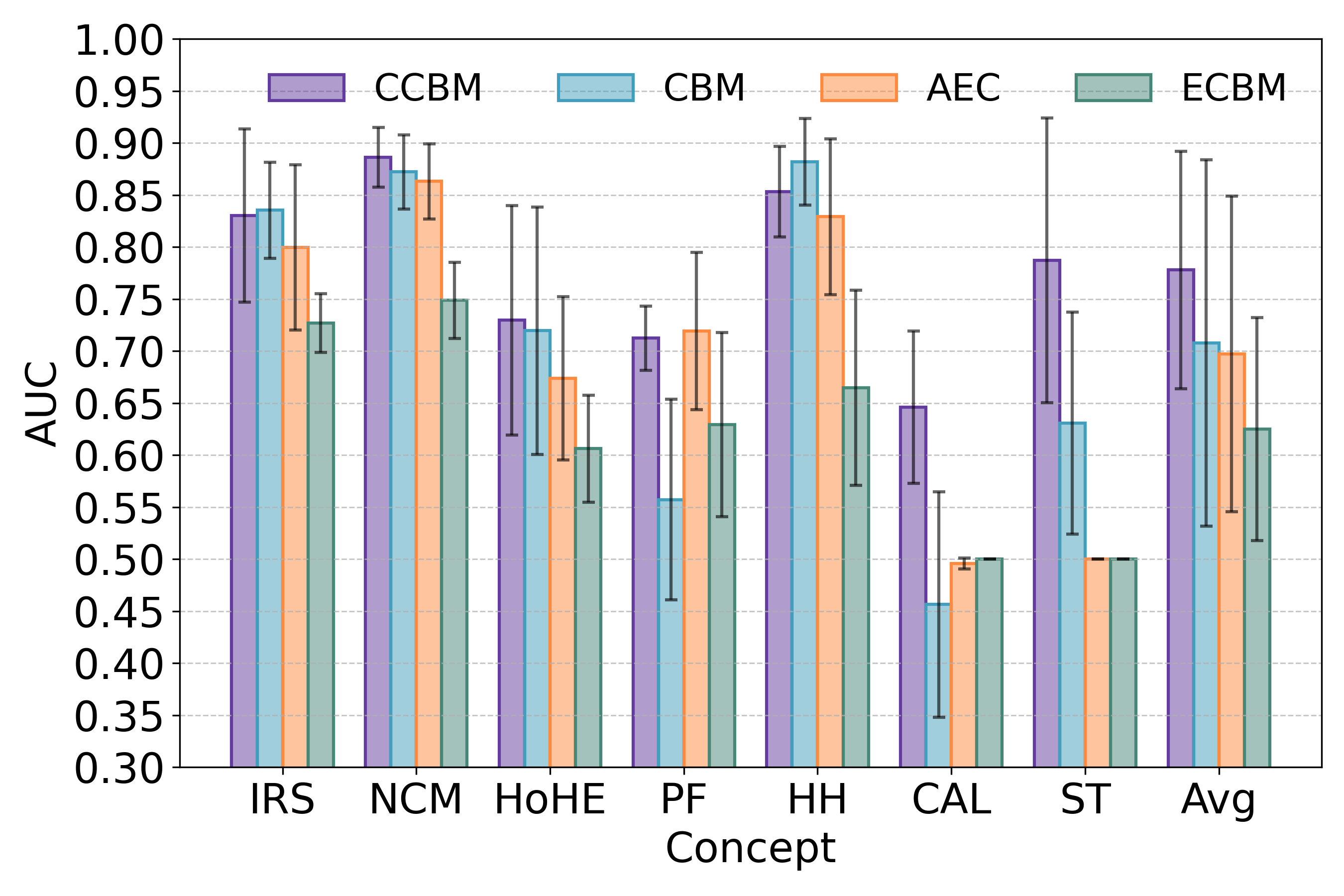}
    \end{minipage}
    \label{fig:cep_auc_BrEaST}
  } \\
  \subfigure[\textit{Concept AUC on Skincon}]{
    \begin{minipage}[b]{1\textwidth}
      \includegraphics[width=1\textwidth]{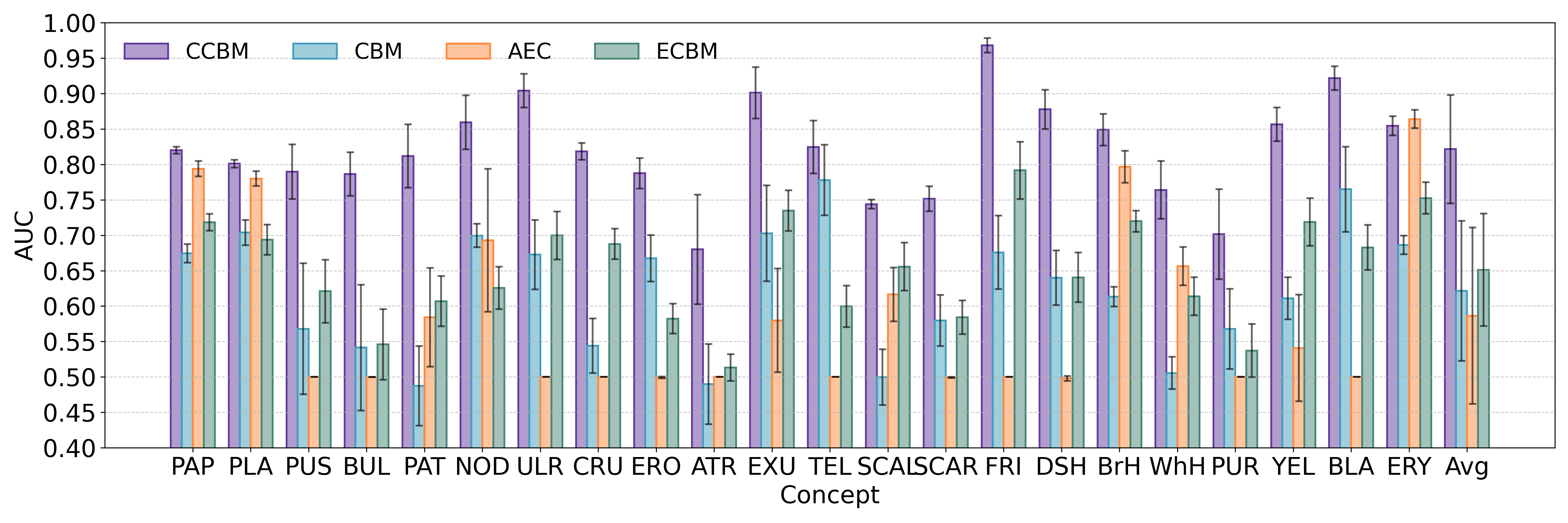}
    \end{minipage}
    \label{fig:cep_auc_skincon}
  } 
  \caption{The fine-grained results of the concept detection task on the \textit{Derm7pt}, \textit{BrEaST} and \textit{Skincon} datasets.The results are the means and stds of five-fold cross-validation experiments. The "Avg" is the mean and variance of AUC over all concepts.} 
  \label{fig:concept_scores}
\end{figure*}

Fig. \ref{fig:concept_scores} displays the fine-grained evaluation results of CCBM and two other competitors on the concept detection task. It is evident that CCBM achieves the highest AUC for most concepts across these three datasets. Particularly, our CCBM demonstrates strong performance across all concepts, while CBM exhibits significant misclassification for certain specific concepts compared to its performance on others, such as the BrH and ErY in the Fig. \ref{fig:cep_auc_skincon}, the PF and CAL concepts in the Fig. \ref{fig:cep_auc_BrEaST}. This indicates that CCBM treats all concepts more equally and provides more accurate concept detection results than other methods. Generally, CCBM has higher average AUC and smaller variance for all concepts. This advantage stems from the concept adaptation mechanism in our model, which enables the extraction of specific concept features for each concept and the calculation of concept scores in their dedicated channels. 

\subsubsection{The Impact of the Number of Unknown Concepts on the Model}

\begin{table*}[htbp]
	\centering
	\caption{Quantitative evaluation of CCBM with different number of unknown concepts. The results are shown as the means $\pm$ stds of five-fold cross-validation experiments. } 
  \scriptsize
	\setlength{\tabcolsep}{2pt} 
	\begin{tabular}{ccccccccc}
	\toprule
	\multicolumn{1}{c}{\multirow{2}{*}{ \makecell{ Dataset \\ (\# of Classes)}}} & \multicolumn{1}{c}{\multirow{2}{*}{$n_u$}} & \multicolumn{3}{c}{Disease Diagnosis} & \multicolumn{2}{c}{Concept Detection} \\
	\cmidrule(lr){3-5} \cmidrule(lr){6-7}
	& & AUC (\%) $\uparrow $ & ACC (\%) $\uparrow $ & F1-score (\%) $\uparrow $ & AUC (\%) $\uparrow $ & ACC (\%) $\uparrow $ \\
	\midrule
	\multicolumn{1}{c}{\multirow{5}{*}{\makecell{\textit{Derm7pt}\\(2)}}} &$0$ & 90.94 $\pm$ 1.84 & 85.85 $\pm$ 1.44 & 83.12 $\pm$ 1.72 & 83.43 $\pm$ 1.19 & 80.95 $\pm$ 1.48\\
	&$1$ & 91.77 $\pm$ 0.42 & 86.46 $\pm$ 1.59 & 83.85 $\pm$ 1.18 & 82.08 $\pm$ 2.03 & \textbf{81.40} $\pm$ \textbf{1.35} \\
	&$2$ & \textbf{93.96} $\pm$ \textbf{0.95} & \textbf{88.15} $\pm$ \textbf{1.64} & \textbf{85.61} $\pm$ \textbf{2.79} & 83.86 $\pm$ 1.00 & 80.90 $\pm$ 1.56 \\
	&$3$ & 92.20 $\pm$ 0.97 & 86.22 $\pm$ 1.97 & 83.35 $\pm$ 2.36 & \textbf{84.48} $\pm$ \textbf{0.94} & 81.24 $\pm$ 0.88 \\
	&$5$ & 93.10 $\pm$ 0.82 & 86.57 $\pm$ 0.93 & 83.37 $\pm$ 1.39 & 84.10 $\pm$ 1.23 & 81.14 $\pm$ 1.57 \\

	\midrule
	\multicolumn{1}{c}{\multirow{5}{*}{\makecell{\textit{Skincon}\\(3)}}} &$0$ & 83.19 $\pm$ 1.52 & 79.35 $\pm$ 1.53 & 65.05 $\pm$ 2.74 & 81.08 $\pm$ 0.42 & 91.27 $\pm$ 0.18 \\
	&$1$ & 84.06 $\pm$ 0.98 & 78.58 $\pm$ 1.61 & 64.11 $\pm$ 1.62 & 82.04 $\pm$ 0.84 & 91.44 $\pm$ 0.84 \\
	&$2$ & 84.11 $\pm$ 1.26 & 79.75 $\pm$ 0.85 & 65.43 $\pm$ 1.50 & \textbf{82.33} $\pm$ \textbf{0.77} & 91.47 $\pm$ 0.10\\
	&$3$ & \textbf{84.55} $\pm$ \textbf{1.87} & \textbf{80.00} $\pm$ \textbf{1.75} & \textbf{67.14} $\pm$ \textbf{1.87} & 82.14 $\pm$ 0.33 & 91.35 $\pm$ 0.25 \\ 
	&$5$ & 83.67 $\pm$ 2.05 & 79.60 $\pm$ 0.97 & 64.49 $\pm$ 1.13 & 81.32 $\pm$ 1.15 & \textbf{91.49} $\pm$ \textbf{0.15} \\

	\midrule
	\multicolumn{1}{c}{\multirow{5}{*}{\makecell{\textit{BrEaST}\\(2)}}} &$0$ & 87.76 $\pm$ 2.89 & \textbf{80.01} $\pm$ \textbf{2.71} & 78.02 $\pm$ 3.08 & 73.19 $\pm$ 2.36 & \textbf{80.56} $\pm$ \textbf{2.50} \\
	&$1$ & 87.39 $\pm$ 5.56 & 79.64 $\pm$ 7.00 & \textbf{78.30} $\pm$ \textbf{7.23} & 74.95 $\pm$ 4.54 & 76.81 $\pm$ 3.86 \\
	&$2$ & \textbf{88.49} $\pm$ \textbf{5.85} & 79.21 $\pm$ 6.00 & 77.95 $\pm$ 5.53 & \textbf{77.78} $\pm$ \textbf{4.00} & 80.06 $\pm$ 3.46 \\
	&$3$ & 86.34 $\pm$ 4.50 & 76.00 $\pm$ 5.07 & 74.01 $\pm$ 4.32 & 74.33 $\pm$ 7.11 & 78.51 $\pm$ 3.16 \\
	&$5$ & 86.94 $\pm$ 4.79 & 78.02 $\pm$ 7.65 & 75.63 $\pm$ 8.94 & 76.21 $\pm$ 3.51 & 80.23 $\pm$ 2.55 \\
	\midrule
	& & AUC (\%) $\uparrow $ & ACC (\%) $\uparrow $ & F1-score (\%) $\uparrow $ & RMSE $\downarrow $ & MAE $\downarrow $ \\
	\midrule
	\multicolumn{1}{c}{\multirow{5}{*}{\makecell{\textit{LIDC-IDRI}\\(2)}}} &$0$ & 90.15 $\pm$ 1.15 & 83.14 $\pm$ 1.73 & 78.97 $\pm$ 2.41 & 0.2366 $\pm$ 0.0045 & 0.1911 $\pm$ 0.0048 \\
	&$1$ & \textbf{90.48} $\pm$ \textbf{0.71} & 75.28 $\pm$ 4.07 & 61.75 $\pm$ 9.31 & 0.2035 $\pm$ 0.0066 & 0.1566 $\pm$ 0.0075 \\
	&$2$ & 90.48 $\pm$ 0.94 & \textbf{84.82} $\pm$ \textbf{0.43} & \textbf{82.56} $\pm$ \textbf{0.67} & \textbf{0.1890} $\pm$ \textbf{0.0046} & \textbf{0.1380} $\pm$ \textbf{0.0032}\\
	&$3$ & 90.21 $\pm$ 1.13 & 84.33 $\pm$ 1.05 & 81.91 $\pm$ 1.62 & 0.1882 $\pm$ 0.0049 & 0.1397 $\pm$ 0.0041 \\
	&$5$ & 90.11 $\pm$ 1.06 & 82.73 $\pm$ 1.41 & 80.63 $\pm$ 1.24 & 0.1918 $\pm$ 0.0042 & 0.1451 $\pm$ 0.0044 \\
	\bottomrule
	\end{tabular}
	\label{tab:Unknown_Concept_Test}
\end{table*}

To further explore the impact of the number of unknown concepts on the CCBM performance, we conduct experiments with different numbers of unknown concepts ($n_u=0,1,2,3,5$). The results are summarized in Table \ref{tab:Unknown_Concept_Test}. CCBM's performance remains competitive with other explainable methods even when the unknown concept learning branch is not enabled. Moreover, the study indicates that setting the number of unknown concepts equal to the number of categories in the datasets generally yields the best performance, although there are slight deviations in some metrics compared to models with other numbers of unknown concepts. 

\subsubsection{Ablation Study}

\begin{table}[t]
	\centering
	\caption{Quantitative evaluation of CCBM with (w) or without (w/o) similarity loss. The results are shown as the means of five-fold cross-validation experiments. } 
  \scriptsize
	\begin{tabular}{ccccccccc}
	\toprule
	\multicolumn{1}{c}{\multirow{2}{*}{Dataset}} & \multicolumn{1}{c}{\multirow{2}{*}{Model}} & \multicolumn{3}{c}{Disease Diagnosis} & \multicolumn{2}{c}{Concept Detection} \\
	\cmidrule(lr){3-5} \cmidrule(lr){6-7}
		& & AUC $\uparrow $ & ACC $\uparrow $ & F1-score $\uparrow $ & AUC $\uparrow $ & ACC $\uparrow $ \\
	\midrule
	\multicolumn{1}{c}{\multirow{2}{*}{\textit{Derm7pt}}} & w & \textbf{93.96} & \textbf{88.15} & \textbf{85.61} & 83.86 & 80.90 \\
	&w/o & 93.78 & 86.70 & 83.80 & \textbf{84.00} & \textbf{80.95} \\
	\midrule
	\multicolumn{1}{c}{\multirow{2}{*}{\textit{Skincon}}} &w & \textbf{84.55} & 80.00 & \textbf{67.14} & 82.14 & 91.35 \\ 
	&w/o & 84.27 & \textbf{80.19} & 66.68 & \textbf{82.28} & \textbf{91.36} \\
	\midrule
	\multicolumn{1}{c}{\multirow{2}{*}{\textit{BrEaST}}} &w & \textbf{88.49} & \textbf{79.21} & \textbf{77.95} & \textbf{77.78} & 80.06 \\
	&w/o & 87.82 & 78.00 & 76.96 & 76.22 & \textbf{80.46} \\
	\midrule
		& & AUC $\uparrow $ & ACC $\uparrow $ & F1-score $\uparrow $ & RMSE $\downarrow $ & MAE $\downarrow $ \\
	\midrule
	\multicolumn{1}{c}{\multirow{2}{*}{\textit{LIDC-IDRI}}} &w & \textbf{90.48} & \textbf{84.82} & \textbf{82.56} & 0.1890 & 0.1380 \\
	&w/o & 90.20 & 84.04 & 81.70 & \textbf{0.1887} & \textbf{0.1378} \\
	\bottomrule
	\end{tabular}
	\label{tab:Abl_study}
\end{table}

The similarity loss $\mathcal{L}_{sim}$ is designed to ensure that the model distinguishes between known and unknown concepts, thereby enhancing performance in disease diagnosis tasks. Essentially, the greater the differentiation between these concepts, the more information the model can assimilate, leading to improved disease diagnosis outcomes. To validate the efficacy of the similarity loss, ablation experiments were performed across the four datasets, with the qualitative results presented in Table \ref{tab:Abl_study}. It is evident that the similarity loss $\mathcal{L}_{sim}$ plays a pivotal role in the model's performance. Models incorporating the similarity loss $\mathcal{L}_{sim}$ outperform those without it across all disease diagnosis metrics, aligning with our expectations. However, models lacking the similarity loss $\mathcal{L}_{sim}$ exhibit superior performance in concept detection tasks across the four datasets. The possible reason is, when concepts are less distinct, the model can leverage information from known concepts more effectively, particularly since these concepts are annotated, thus benefiting the concept detection.

\subsection{Explanability Analysis}
Following the prior research \cite{koh2020concept, bie2024mica}, we analyze the explanability of our model across the dimensions of faithfulness, efficiency and plausibility. Furthermore, we qualitatively evaluate the effectiveness of our learned unknown concepts by comparing the explanations of model with $n_u=0$ and $n_u=c$.

\subsubsection{Inference-time Intervention for Faithfulness}
\begin{figure}[htbp]
	\centering
	\subfigure[\textit{Derm7pt}]{
		\begin{minipage}[b]{0.22\textwidth}
			\includegraphics[width=1\textwidth]{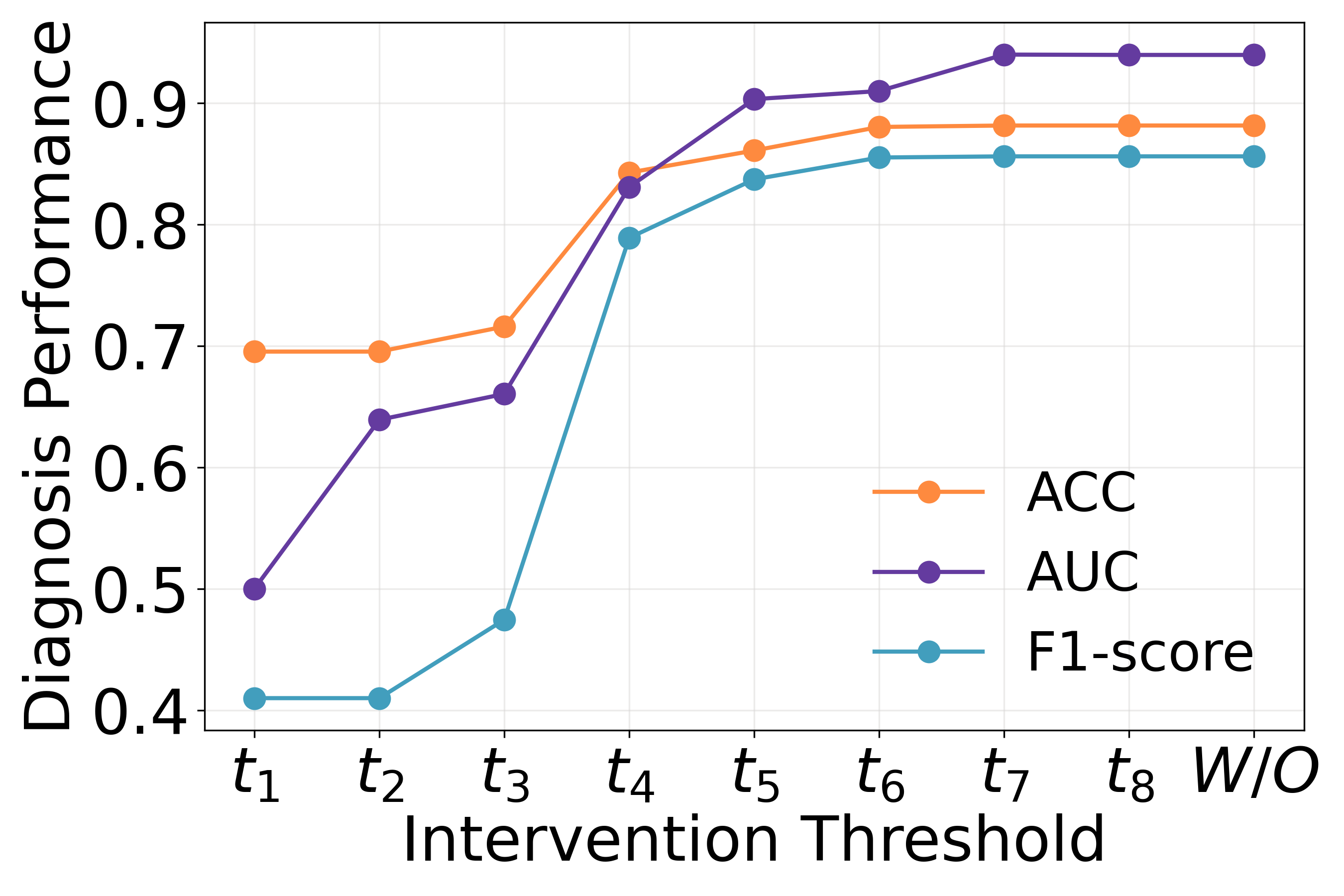}
		\end{minipage}
		\label{fig:Intervention_Derm7pt}
	} 
	\subfigure[\textit{Skincon}]{
			\begin{minipage}[b]{0.22\textwidth}
			\includegraphics[width=1\textwidth]{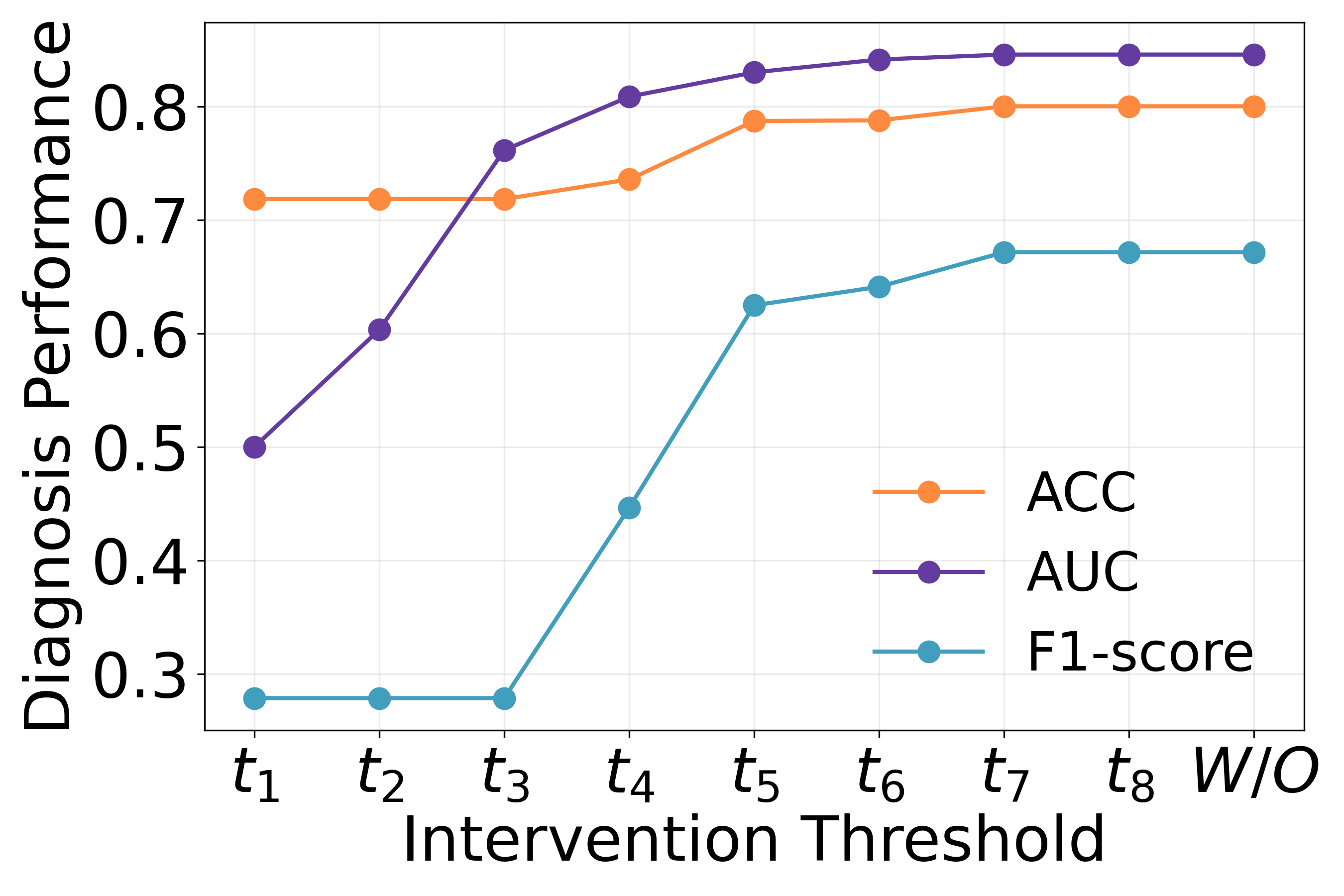}
			\end{minipage}
		\label{fig:Intervention_Skincon}
	 }
	\subfigure[\textit{BrEaST}]{
			\begin{minipage}[b]{0.22\textwidth}
			\includegraphics[width=1\textwidth]{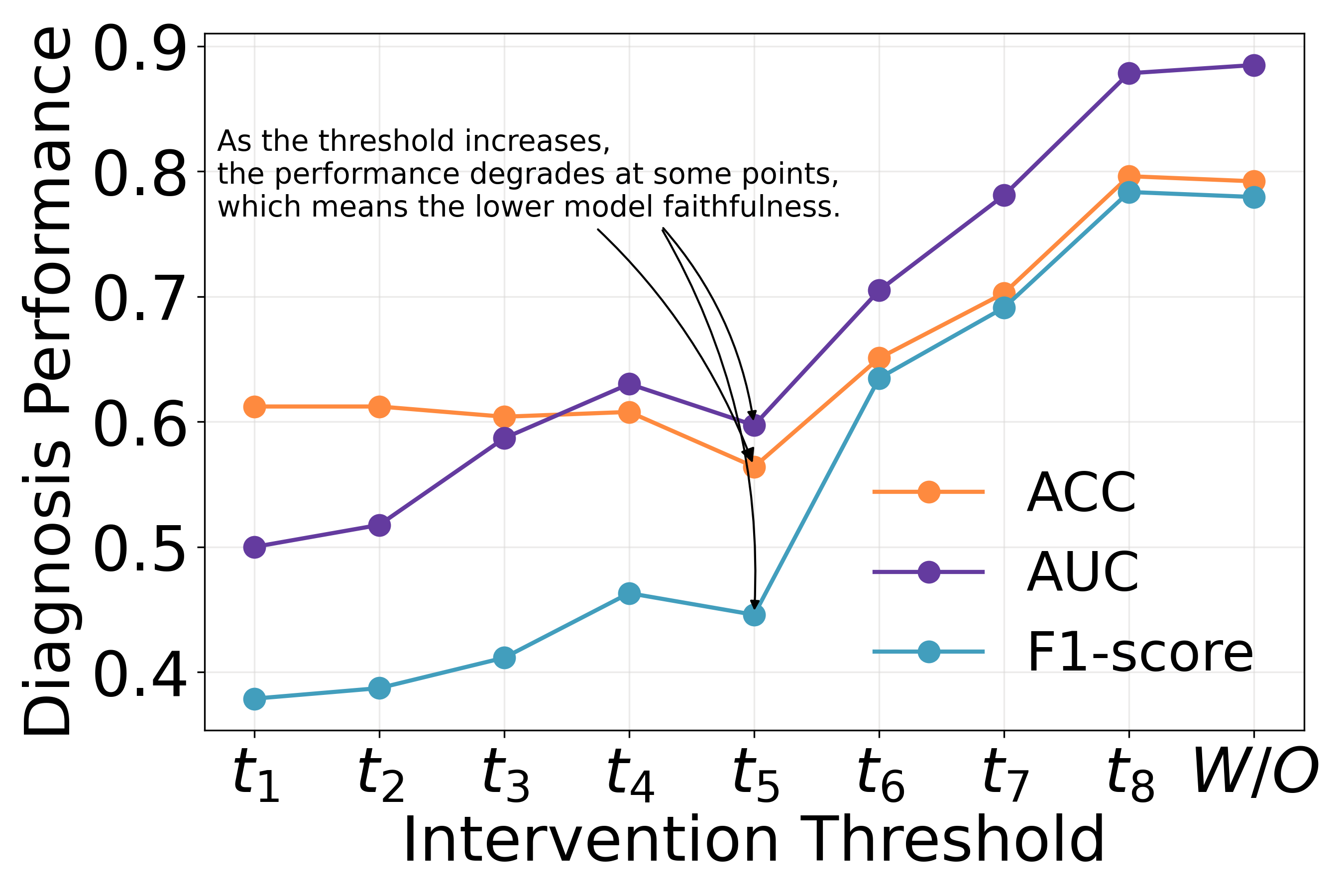}
			\end{minipage}
		\label{fig:Intervention_BrEaST}
	}
	\subfigure[\textit{LIDC-IDRI}]{
			\begin{minipage}[b]{0.22\textwidth}
			\includegraphics[width=1\textwidth]{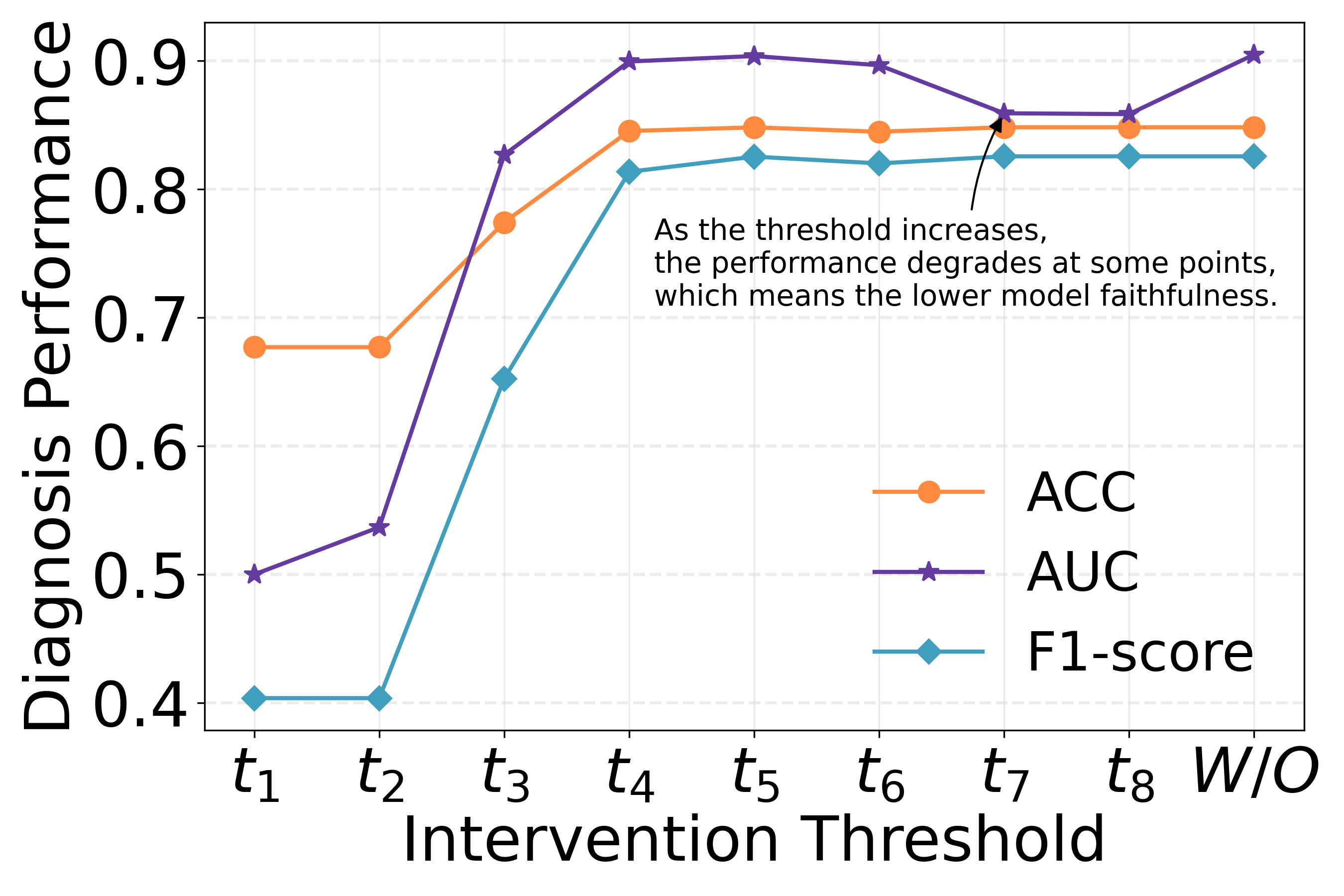}
			\end{minipage}
		\label{fig:Intervention_LIDC}
	}
	\caption{Inference-time intervention results. The $x$-axis represents the thresholds ($t_1 \leq t_2 \leq ... \leq t_8$), and the $y$-axis represents the diagnosis performance after intervention. }
	\label{fig:Intervention}
	\end{figure}

Faithfulness indicates that the model explanations could faithfully elucidate the model decision. In the inference-time intervention experiments, we artificially modify partial concept predictions to observe the resulting changes in the final model decisions, allowing us to assess the effectiveness of the concept explanations. Specifically, we establish a set of thresholds for concept scores during model inference, where any concept scores surpassing the threshold are reset to zero. Subsequently, we present the diagnosis outcomes inferred using the intervened concept scores. Fig. \ref{fig:Intervention} shows that the disease diagnosis performance is notably impacted by the intervention of concept scores. For these four datasets, the AUC, ACC and F1-score generally exhibit significant improvements as the intervention threshold increases. When the threshold is set too low, the model performance deteriorates (with AUC dropping to 50\%). These intervention experiments underscore the model's heavy reliance on predicted concept scores and affirm the faithfulness of the explanations provided. However, fluctuations in trends are observed for the \textit{BrEaST} dataset and the \textit{LIDC-IDRI} with large thresholds. In the case of the \textit{BrEaST} dataset, this variability might be attributed to the dataset's relatively small size, hindering robust concept learning. Conversely, for \textit{LIDC-IDRI} dataset, the regression nature of the concept learning task in LIDC-IDRI presents greater challenges in aiding model decisions compared to concept classification. 

\subsubsection{Label Efficiency}

\begin{figure}[t]
	\centering
	\subfigure[\textit{Derm7pt}]{
		\begin{minipage}[b]{0.22\textwidth}
			\includegraphics[width=1\textwidth]{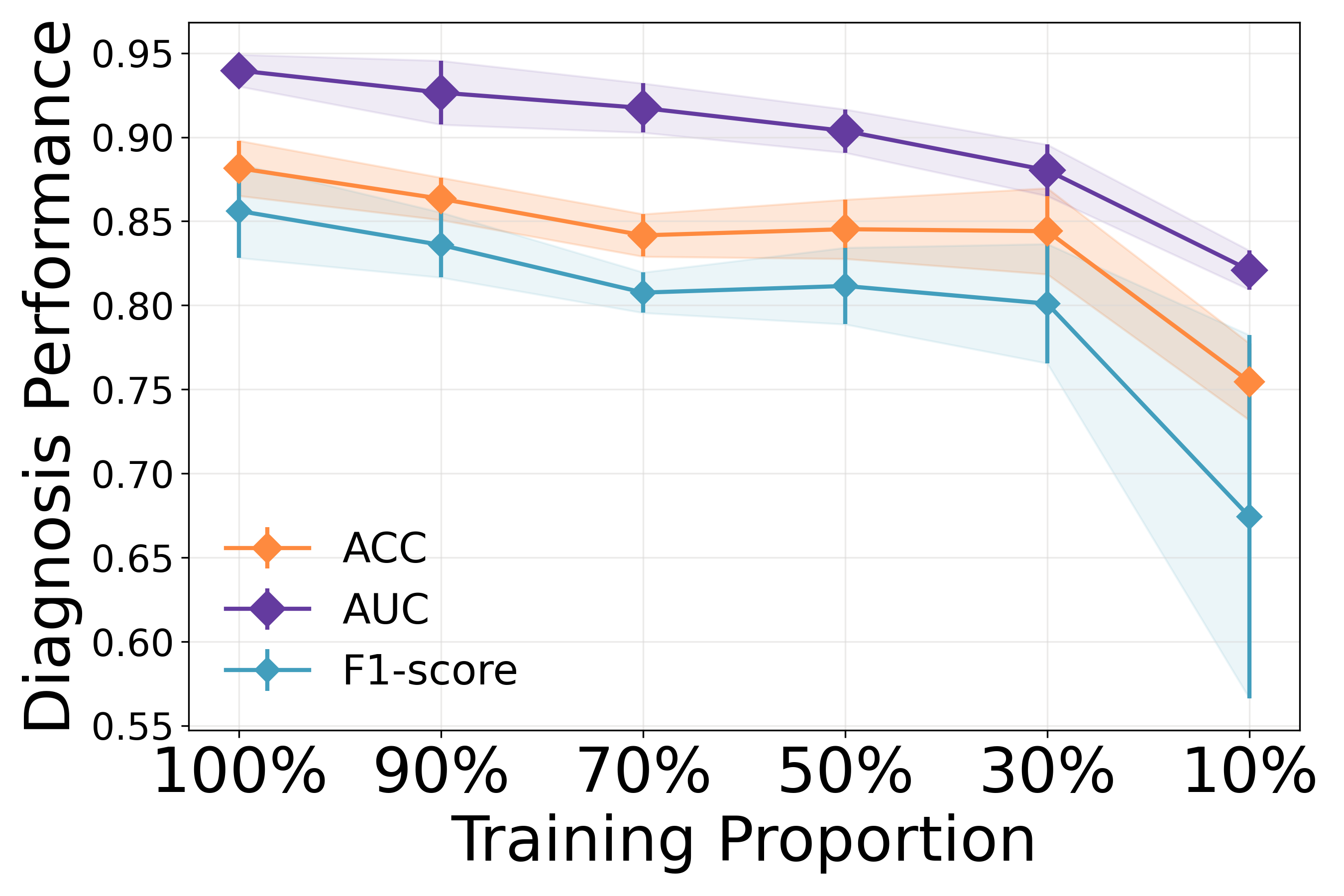}
		\end{minipage}
		\label{fig:effic_Derm7pt}
	} 
	\subfigure[\textit{Skincon}]{
			\begin{minipage}[b]{0.22\textwidth}
			\includegraphics[width=1\textwidth]{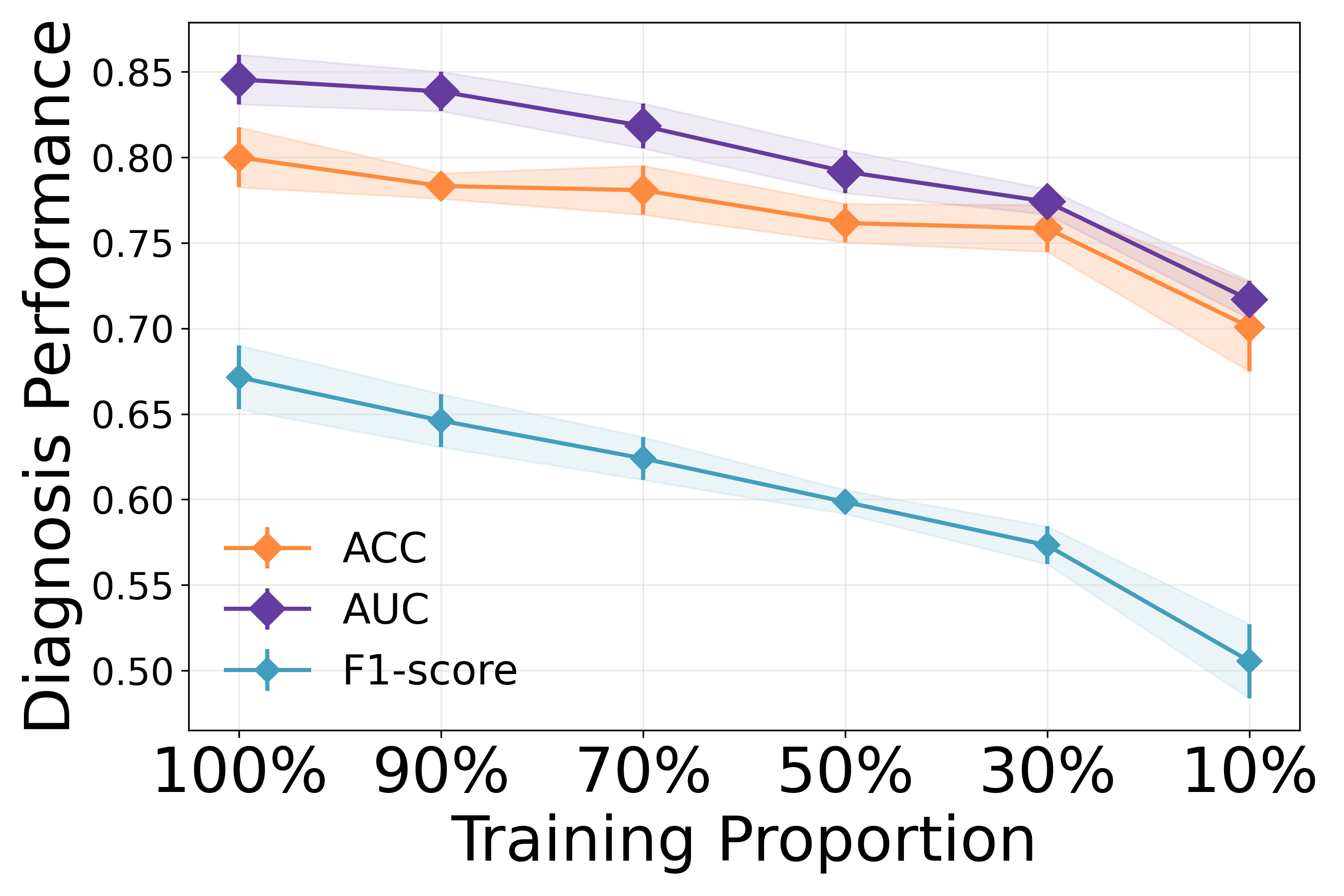}
			\end{minipage}
		\label{fig:effic_Skincon}
	 }
	\subfigure[\textit{BrEaST}]{
			\begin{minipage}[b]{0.22\textwidth}
			\includegraphics[width=1\textwidth]{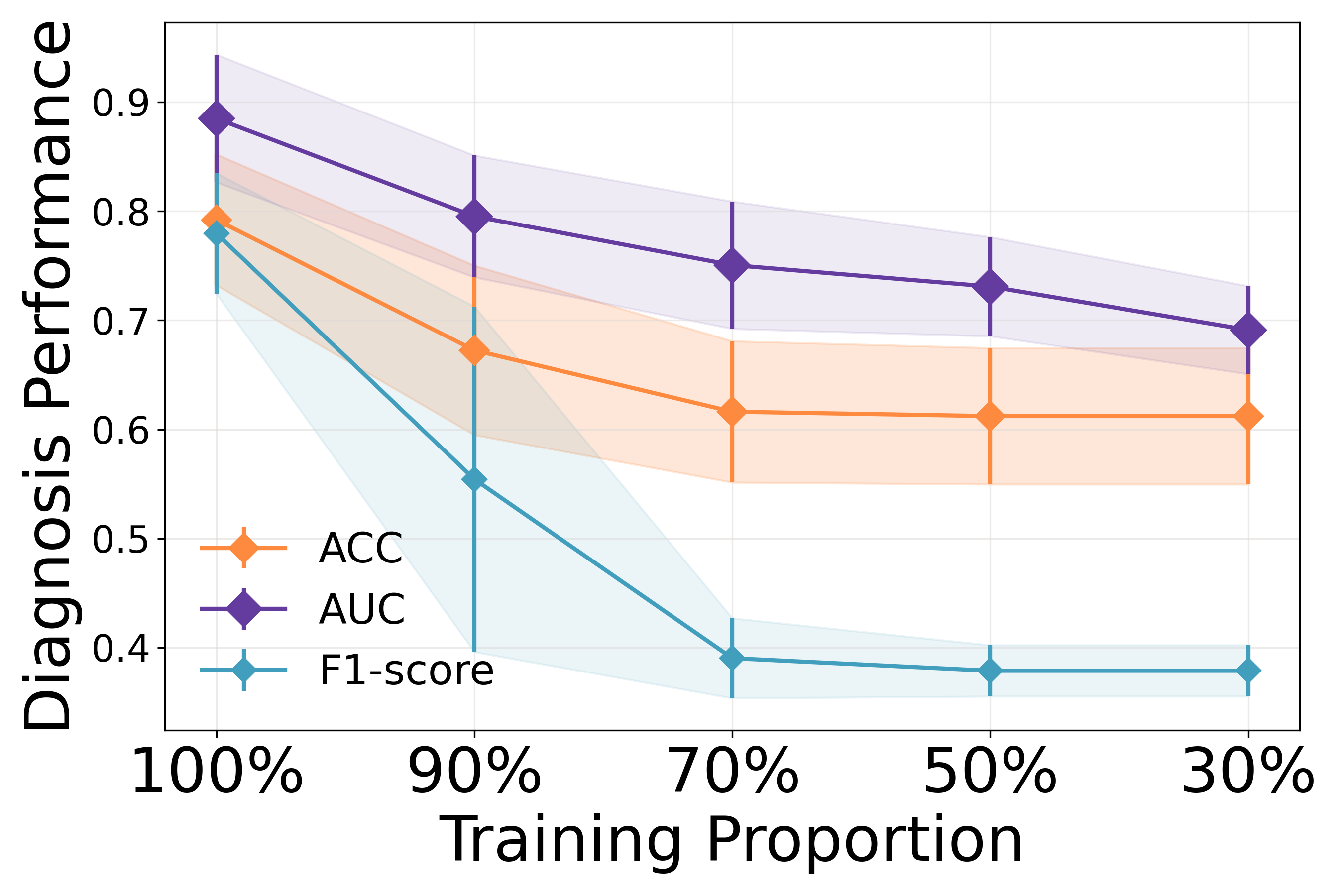}
			\end{minipage}
		\label{fig:effic_BrEaST}
	}
	\subfigure[\textit{LIDC-IDRI}]{
			\begin{minipage}[b]{0.22\textwidth}
			\includegraphics[width=1\textwidth]{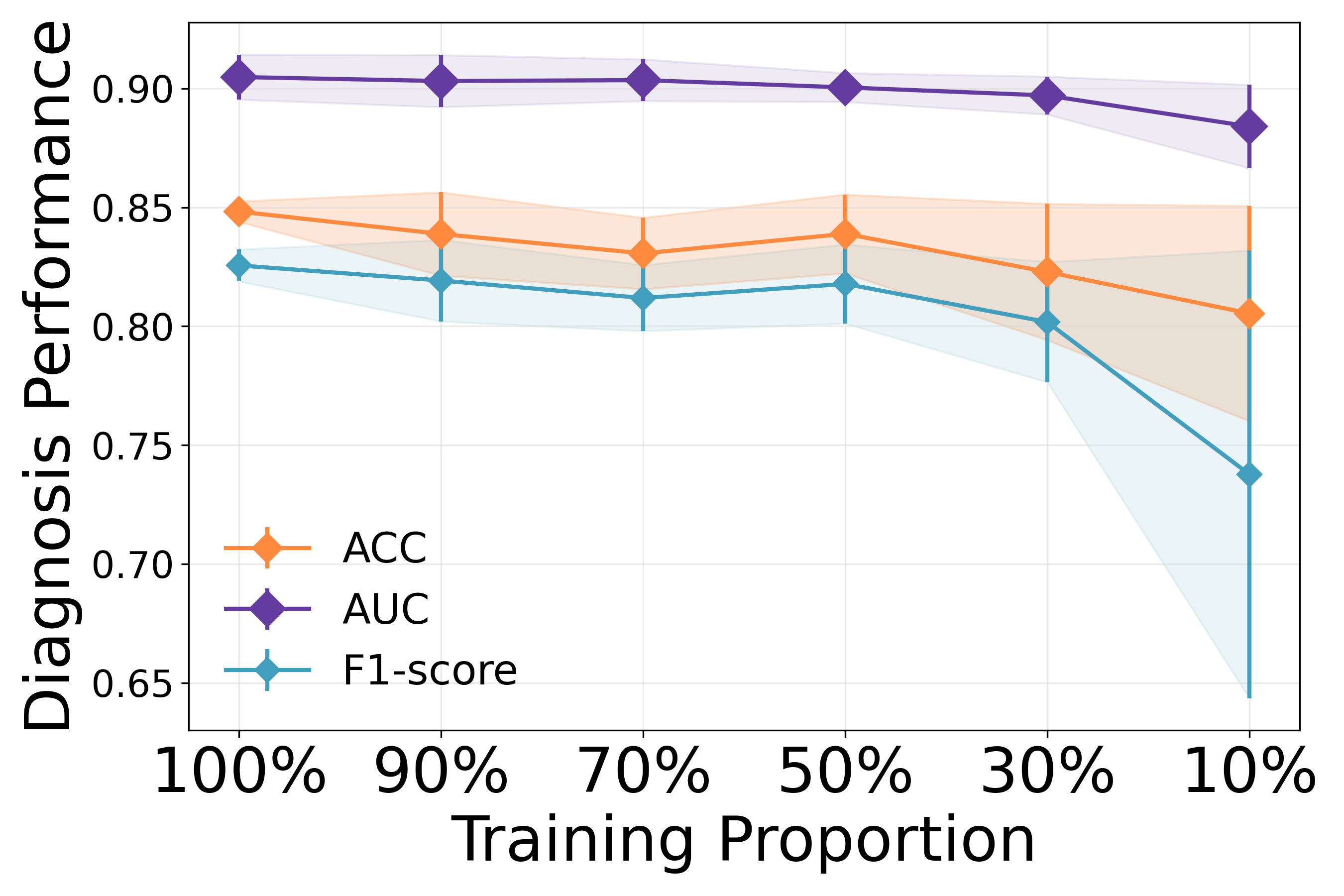}
			\end{minipage}
		\label{fig:effic_LIDC}
	}
	\caption{Label efficiency experiment results. The $x$-axis and $y$-axis represent the training proportion and diagnosis performance, respectively. }
	\label{fig:effic}
	\end{figure}

Obtaining diagnosis labels and annotations for medical images, crucial for improving model interpretability, is often challenging. Hence, ensuring a strong correlation between concepts and model decisions is essential for leveraging labeled data effectively. To validate the label effectiveness of CCBM, we vary the proportions of fully annotated training data in each five-fold cross-validation experiment while retaining the original full test data for testing. As depicted in Fig. \ref{fig:effic}, the model performance declines as the Training Data Proportion (TDP) decreases, albeit with only slight drops in the three metrics. Across the \textit{Derm7pt}, \textit{Skincon}, and \textit{LIDC-IDRI} datasets, model performance decreases gradually, with metrics declining by 5\% to 10\%. The \textit{BrEaST} dataset exhibits a more pronounced performance decline, attributed to its small training data size. On the \textit{LIDC-IDRI} dataset, performance degradation remains modest until TDP decreases to 30\%. However, a significant performance drop occurs when TDP reaches 10\%, deviating from previous trends. These experiments shows CCBM's efficient utilization of concept and diagnostic labels during training and CCBM demonstrates robust performance when the number of well-annotated samples exceeds approximately 200.

\subsubsection{Visual and Textual Explanations for Plausibility} 

	\begin{figure*}[t]
	\begin{center}
		\includegraphics[width=1\linewidth]{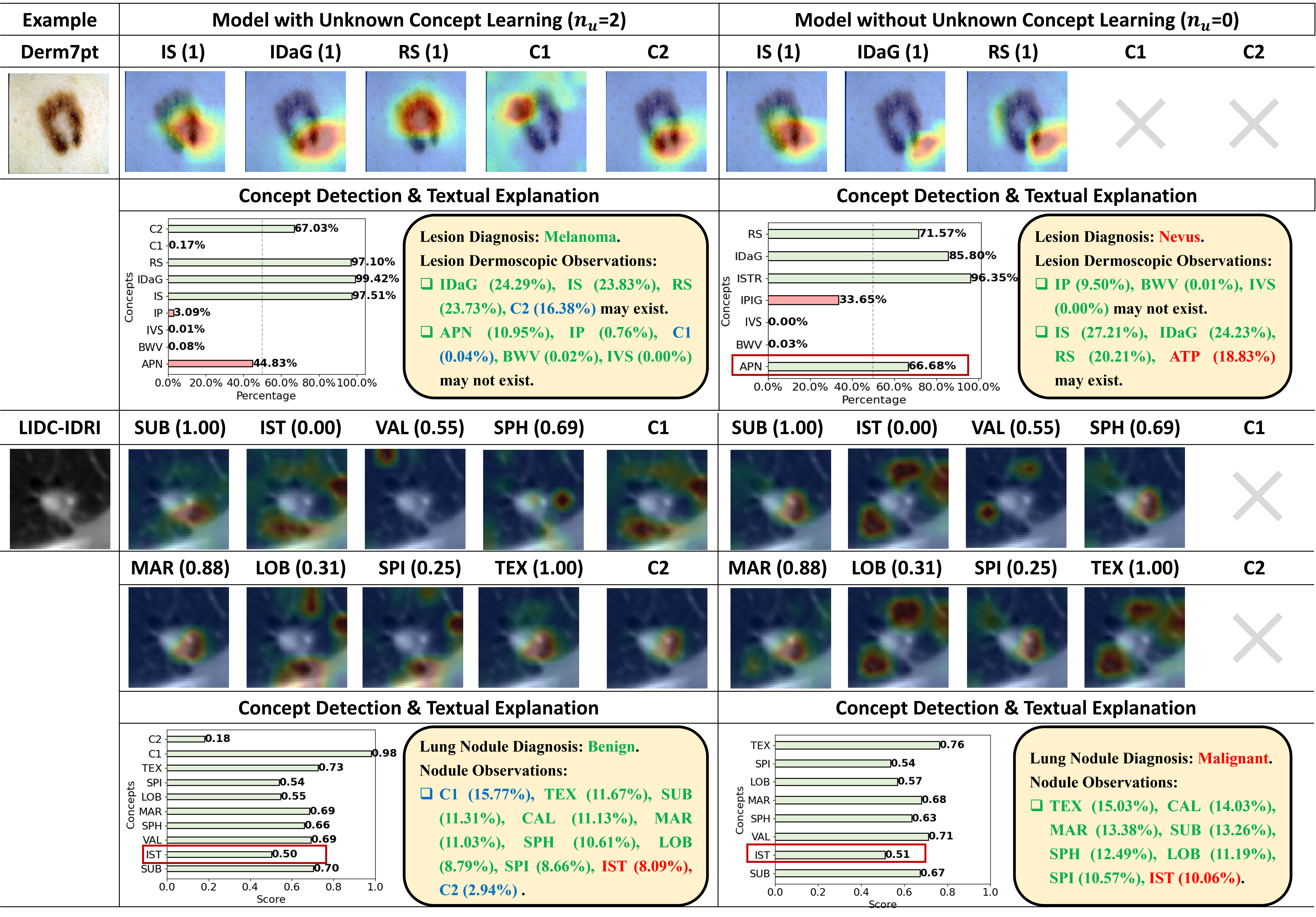}
	 \end{center}
	\caption{Visual and textual explanations of two images from \textit{Derm7pt} and \textit{LIDC-IDRI}, respectively. We visualize the known concepts and unknown concepts (named C1 and C2) of these two examples. The value in brackets represents the truth label of the concept. (For \textit{Derm7pt}, only the existing concepts are shown. ) Concept scores in red box indicate incorrectly predicted concepts. In the textual explanations, green text denotes correct predictions, while red text highlights incorrect predictions. Blue text presents the insights from the learned unknown concepts. }
	\label{fig: explanation}
	\end{figure*}

The plausibility of explanations is a critical issue in explainable artificial intelligence, encompassing the necessity for model explanations to be both convincing in real-world scenarios and understandable to humans. We present local visual explanations and textual insights regarding the decision-making process of our CCBM model. For visual explanations, we generate concept activation maps using Grad-CAM and visualize the model's concept scores. Additionally, we include textual summaries of the final disease diagnosis results and the contributions of all concepts. In Fig. \ref{fig: explanation}, we showcase the explanations for two images from \textit{Derm7pt} and \textit{LIDC-IDRI} datasets. Specifically, we select images where the model, without the setting for learning unknown concepts, makes incorrect predictions, but the model equipped with the unknown concept branch accurately diagnoses these cases. This demonstration aims to highlight the effectiveness of the learned unknown concepts.

From the observed cases, we can find that CCBM with the unknown concept learning setting offers accurate diagnoses by leveraging additional insights from unknown concepts. When comparing the concept scores between the model without the unknown concept learning setting and the one with it, the latter provides more precise concept scores that closely align with the ground truth for the correctly predicted concepts. The textual explanations further elucidate the influence of the learned unknown concepts, aiding in making accurate diagnoses. These results underscore that CCBM exhibits enhanced credibility and comprehensibility in terms of interpretability.

\section{Conclusion}

In this paper, we propose a concept complement bottleneck model for interpretable medical image analysis by jointly learning unknown concepts while using known concepts to predict disease. Our model incorporates concept adapters and aggregators with a visual-text concept cross-attention module, creating a fairier concept bottleneck model that enhances the precision and effectiveness of disease predictions using known concepts. We also present an effective strategy for learning unknown concepts, aiming at extracting more significant information to enhance model performance. Through comprehensive experiments, we demonstrate that our model achieves superior classification performance in concept detection and disease diagnosis tasks, providing more faithful and understandable explanations. Detailed experiment analysis showcases the effectiveness of the learned unknown concepts. Our study has not yet explored the textualization, specialization, and generalization of the discovered new concepts, which need further exploration in the future.

\section*{Declaration of competing interest} 
The authors declare that they have no known competing financial interests or personal relationships that could have appeared to influence the work reported in this paper.

\section*{Data availability}
Data availability Data will be made available on request.

\section*{Acknowledgements}
This work was supported by the Hong Kong Innovation and Technology Fund (Project No. MHP/002/22), HKUST (Project No. FS111), and the Research Grants Council of the Hong Kong Special Administrative Region, China (Project Reference Number: T45-401/22-N). 




\bibliographystyle{elsarticle-num-names} 
\bibliography{CCBM}







\end{document}